\documentclass[sigplan,screen]{acmart}
\AtBeginDocument{%
  }

\copyrightyear{2025}
\acmYear{2025}
\setcopyright{acmlicensed}\acmConference[ASPLOS '25]{Proceedings of the 30th ACM International Conference on Architectural Support for Programming Languages and Operating Systems, Volume 2}{March 30-April 3, 2025}{Rotterdam, Netherlands}
\acmBooktitle{Proceedings of the 30th ACM International Conference on Architectural Support for Programming Languages and Operating Systems, Volume 2 (ASPLOS '25), March 30-April 3, 2025, Rotterdam, Netherlands}
\acmDOI{10.1145/3676641.3716269}
\acmISBN{979-8-4007-1079-7/2025/03}

\usepackage[linesnumbered,ruled,vlined]{algorithm2e}
\usepackage{multirow}
\usepackage{graphicx}
\usepackage{mathtools}
\usepackage{balance}
\usepackage{float}

\begin{document}

\title{Pruner: A Draft-then-Verify Exploration Mechanism to Accelerate Tensor Program Tuning}

\author{Liang Qiao}
\authornote{Both authors contributed equally to this research.}
\affiliation{%
  \institution{University of Science and Technology of China}
  \city{Hefei}
  \country{China}}
\email{ql1an9@mail.ustc.edu.cn}

\author{Jun Shi}
\authornotemark[1]
\affiliation{%
  \institution{University of Science and Technology of China}
  \city{Hefei}
  \country{China}}
\email{shijun18@ustc.edu.cn}

\author{Xiaoyu Hao}
\affiliation{%
  \institution{University of Science and Technology of China}
  \city{Hefei}
  \country{China}
}
\email{hxy2018@mail.ustc.edu.cn}

\author{Xi Fang}
\affiliation{%
  \institution{University of Science and Technology of China}
  \city{Hefei}
  \country{China}
}
\email{fangxi@mail.ustc.edu.cn}

\author{Sen Zhang}
\affiliation{%
  \institution{University of Science and Technology of China}
  \city{Hefei}
  \country{China}
}
\email{sen@mail.ustc.edu.cn}

\author{Minfan Zhao}
\affiliation{%
  \institution{University of Science and Technology of China}
  \city{Hefei}
  \country{China}
}
\email{zmf@mail.ustc.edu.cn}

\author{Ziqi Zhu}
\affiliation{%
  \institution{University of Science and Technology of China}
  \city{Hefei}
  \country{China}
}
\email{ta1ly@mail.ustc.edu.cn}

\author{Junshi Chen}
\authornote{Also with Laoshan Laboratory, Qingdao, China}
\affiliation{%
  \institution{University of Science and Technology of China}
  \city{Hefei}
  \country{China}
}
\email{cjuns@ustc.edu.cn}

\author{Hong An}
\authornote{Corresponding author.}
\authornotemark[2]
\affiliation{%
  \institution{University of Science and Technology of China}
  \city{Hefei}
  \country{China}
}
\email{han@ustc.edu.cn}

\author{Xulong Tang}
\affiliation{%
  \institution{University of Pittsburgh}
  \city{Pittsburgh}
  \country{USA}
}
\email{tax6@pitt.edu}

\author{Bing Li}
\affiliation{%
  \institution{NIO}
  \city{Shanghai}
  \country{China}
}
\email{libing475211023@sjtu.edu.cn}

 \author{Honghui Yuan}
 \email{ethan.song@nio.com}
\author{Xinyang Wang}
 \email{12307130155@fudan.edu.cn}
 \affiliation{%
  \institution{NIO}
  \city{Shanghai}
  \country{China}
}

% \author{Honghui Yuan}
% \affiliation{%
%   \institution{NIO}
%   \city{Shanghai}
%   \country{China}
% }
% \email{ethan.song@nio.com}

% \author{Xinyang Wang}
% \affiliation{%
%   \institution{NIO}
%   \city{Shanghai}
%   \country{China}
% }
% \email{12307130155@fudan.edu.cn}

\renewcommand{\shortauthors}{Liang Qiao et al.}

\begin{abstract}

Tensor program tuning is essential for the efficient deployment of deep neural networks. Search-based approaches have demonstrated scalability and effectiveness in automatically finding high-performance programs for specific hardware. However, the search process is often inefficient, taking hours or even days to discover optimal programs due to the exploration mechanisms guided by an accurate but slow-learned cost model. Meanwhile, the learned cost model trained on one platform cannot seamlessly adapt online to another, which we call cross-platform online unawareness.

In this work, we propose  Pruner and MoA-Pruner.  Pruner is a "Draft-then-Verify" exploration mechanism that accelerates the schedule search process.  Instead of applying the complex learned cost model to all explored candidates,   Pruner drafts small-scale potential candidates by introducing a naive Symbol-based Analyzer (draft model), then identifies the best candidates by the learned cost model.  MoA-Pruner introduces a Momentum online Adaptation strategy to address the cross-platform online unawareness.

We incorporate Pruner into the TVM and conduct extensive experiments on three GPU-based platforms.  Results show considerable speedup in schedule search time. In online tuning scenarios, Pruner and MoA-Pruner achieve an average speedup of $2.6 \times$ and $4.82 \times$ compared to Ansor. In offline tuning scenarios, Pruner achieves an average speedup of $4.75 \times$ and $4.05\times$ compared to TenSet and TLP, respectively.  Furthermore, Pruner achieves an average speedup of $4.08 \times$ compared to MetaSchedule on TensorCore.

\end{abstract}




% \received{24 June 2024}
% \received[revised]{3 November 2024}
% \received[accepted]{28 January 2025}

\maketitle

\section{Introduction}

Deep learning accelerators (DLAs) have promoted the widespread application of deep neural networks (DNNs) in various domains, such as autonomous driving, augmented reality, etc. To accelerate model inference, tensor program optimization has emerged as a critical process to maximize hardware computing efficiency.   Existing deep learning frameworks \cite{paszke2019pytorch, abadi2016tensorflow} express the DNNs as a computation graph, in which nodes represent the operators (e.g., convolution, matrix multiplication), and map these operators onto manually optimized kernel libraries (e.g., cuDNN, MKL-DNN) for specific DLAs. However, the manual tuning of these kernel libraries for each DLA and operator requires significant expertise and effort. This intensive manual effort hinders the efficient development and innovation of new operators and custom-designed DLAs.  Search-based deep learning compilers (DLCs) \cite{kjolstad2017tensor,chen2018tvm, xla,ma2020rammer} are recent and more scalable approaches to automate the tensor programs tuning, improving the deployment efficiency of DNNs on various DLAs. Given an operator,  search-based DLCs \cite{chen2018tvm, chen2018learning, zheng2020ansor} typically define a search space of schedules and search for the best-compiled tensor programs tailored towards different DLAs.  The core of the search process is the cost model, which estimates the performance of tensor program candidates to reduce the time-consuming on-device measurements.

With the growing interest in using deep learning techniques to learn a cost model \cite{baghdadi2021deep, zheng2021tenset, zhao2024felix, zhai2023tlp, kaufman2021learned, steiner2021value}, designing efficient search-based DLCs suffers from the following challenges. First,  fully relying on a learned cost model makes the search extremely time-consuming. Search-based DLCs use predicted latency from the learned cost model as the metric \cite{zheng2020ansor, zhai2023tlp, shao2022tensor, zhao2024felix} to search in a tuning space,  which is determined by various tunable variables (e.g., tile sizes, unroll factors),  and the size of this combinatorial space usually reaches billions on GPUs \cite{zheng2020ansor}. However, applying the learned cost model to all candidates during the search process is more expensive than using the empirical formula cost model \cite{mullapudi2016automatically} due to the complex CPU-based feature extraction and GPU-based cost model inference. Second, feature engineering determines the upper bound of the learned cost model's performance. It involves encoding tensor programs into features that allow the model to learn the relationship between these features and performance accurately. For instance, Ansor \cite{zheng2020ansor} and TenSet \cite{zheng2021tenset} manually extracted 164-dimensional features for each innermost non-loop statement in the context of a full program and used a Multi-Layer Perceptron (MLP) model, whereas TIRAMISU \cite{baghdadi2021deep} manually extracted 2534-dimensional features and utilized an LSTM model. Despite its importance, feature engineering requires domain experts with deep knowledge of hardware architecture, making the process labor-intensive and complex. Finally, effectively adapting a cross-platform pre-trained cost model to help train a target cost model presents significant challenges. The domain gap between platforms often leads to notable performance variations for an operator, even when using the same tuning configurations (e.g., tile size) across different platforms. This variation poses a problem: a cost model well-trained on one platform typically cannot be applied to another in online cost model tuning, which we call cross-platform online unawareness.

Prior works fall short of solving all these challenges in a comprehensive manner.  First, to accelerate the search process, existing approaches, like constraint-based genetic algorithm \cite{bi2023heron} and gradient-based exploration mechanism \cite{zhao2024felix}, introduce additional constraints for guiding the exploration direction and reduce the exploration iteration. While TLP \cite{zhai2023tlp} reduces the time overhead associated with feature engineering.  However, these methods still apply complex cost models to each schedule candidate during the search process, resulting in the significant inference overhead of the cost model. Second,  to optimize feature engineering, some works have attempted to design features that are easy to extract, such as TLP, which extracts features from high-level schedule primitives and introduces a Transformer-based cost model to learn the temporal relations. However, this design presents challenges for model training, requiring extensive offline pre-training data (e.g., the TenSet dataset) to achieve high accuracy. Constructing tensor program datasets for all platforms is time-consuming (e.g., over 10 days for a small dataset with approximately 1.5 million tensor programs in our experiments). This limitation reduces TLP's utility for online tuning scenarios like Ansor. Finally, to address cross-platform online unawareness, online finetuning is a method, but the limited and biased data collected during the early stages of online training can disrupt model training. TenSet \cite{zheng2021tenset} employs transfer learning and trains a local model to predict the gap between the cross-platform model and target hardware. However, despite using the same amount of online collected data, the training complexity of the local model remains essentially equivalent to directly training a target model from scratch. Moses \cite{zhao2022moses} utilizes model distillation, which necessitates additional evaluation of the transferability of each weight. However, these methods do not significantly alleviate the challenges associated with online model cross-platform adaptation. TLP uses multi-task learning, still requiring the construction of a dataset tailored to the target platform and does not support online adaptation. Therefore, none of the existing works effectively address all three challenges simultaneously, which is our focus.

In this paper, we propose Pruner and MoA-Pruner.  Pruner is a "Draft-then-Verify" exploration mechanism that accelerates the schedule search process.  Pruner has two key components for "Draft" and "Verify", respectively. First,  Pruner introduces a latent schedule explorer (\textbf{LSE} as "Draft") that treats the exploration object as a maximized hardware fitness problem by using a naive Symbol-based Analyzer (draft model) and drafts small-scale candidates from all explored candidates as an input for the learned cost model.  Second, Pruner designs a pattern-aware cost model  (\textbf{PaCM} as "Verify") to explore temporal dataflow patterns aligned with program behaviors. PaCM verifies and identifies the best candidates from drafted small-scale candidates instead of applying the complex learned cost model to all explored candidates.  MoA-Pruner introduces the momentum online adaptation, a strategy that succeeded in self-supervised learning \cite{he2020momentum, chen2020improved},  to enable efficient online adaptation to any platform. MoA-Pruner treats the cross-platform pre-trained model as a Siamese model, initializing the target model with its weights and continuously updating itself for adaptation to the target platform.

In our experiments,  we validated Pruner's feasibility and efficiency on three GPU-based platforms. With reaching the performance of other approaches tuning 2,000 trials, Pruner and MoA-Pruner can achieve an average speedup of $2.6 \times$ and $4.82 \times$ compared to Ansor in online tuning scenarios. In offline tuning scenarios, Pruner achieves an average speedup of $4.75 \times$ and $4.05 \times$ compared to TenSet and TLP, respectively.  Furthermore, Pruner achieves an average speedup of $4.08 \times$ compared to MetaSchedule on TensorCore. Notably, MoA-Pruner applies to all automatic search frameworks that rely on space exploration with a learned cost model.

Our main contributions can be summarized as follows:

\begin{itemize}
\item  We propose Pruner, a "Draft-then-Verify" exploration mechanism for rapid and high-quality schedule space exploration.

\item We propose a latent schedule explorer that performs hardware-aware symbolic analysis for draft generation and a pattern-aware cost model that captures the critical temporal dataflow features for performance verification.

\item We introduce MoA-Pruner, using a momentum online adaptation strategy to address the cross-platform online unawareness in online cost model tuning scenarios.

\item  We incorporate these techniques into the TVM and conduct comprehensive evaluations. The results show that the proposed methods outperform the state-of-the-art approaches on various DNNs with significantly reduced search time.
\end{itemize}

Code is available at \url{https://github.com/qiaolian9/Pruner}.

\section{Background}

\subsection{Search-based deep learning compilers} 
\begin{figure}[t]
  \centering
  \includegraphics[width=1\columnwidth]{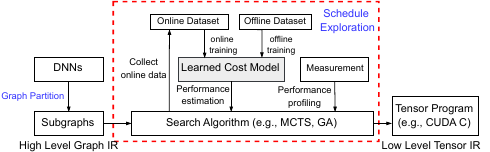}
  \caption{The workflow of search-based DLCs. The red dashed box is the optimization workspace of the Pruner.}
  \label{naivetuningpipeline}
  \Description{}
\end{figure}

Figure \ref{naivetuningpipeline} shows the typical workflow of common search-based DLCs with a learned cost model such as TVM \cite{shao2022tensor, zheng2020ansor, chen2018tvm}, Halide \cite{adams2019learning, anderson2020learning}, etc. Those compilers accept the DNN or a computational graph in high-level mathematical expression as input and then divide the corresponding computational graph into multiple subgraphs through several computational graph optimizations. Each subgraph has its own search space, typically determined by various tunable variables (e.g., tile sizes, unroll factors) in schedule primitives. In each standard tuning round, these compilers apply search algorithms, including genetic algorithm (GA), beam search, Monte Carlo tree search (MCTS), etc,  to explore search space and find optimal schedules. By exploring an extensive range of optimization combinations, these compilers can frequently discover programs that surpass hand-optimized implementations. Due to the huge size of the search space and the time-consuming on-device measurement, it is impossible to measure the execution time of every program candidate. Therefore, it uses a learned cost model as a search guidance metric and selects the best-predicted candidates to measure them on the target platform to find the optimal tensor program. Meanwhile, update the learned cost mode in online tuning scenarios. Finally, these high-performance tensor programs are delivered to specific accelerated backends, such as CUDA, to generate the final executable.

\subsection{Learned cost models and cross-platform transfer} 
Many learned cost models have been proposed \cite{zheng2021tenset, zhai2023tlp, adams2019learning, baghdadi2021deep, kaufman2021learned, zhao2024felix}. Some works \cite{zheng2021tenset, zhao2024felix} began using simple deep learning models such as MLP and outperformed those \cite{chen2018learning, zheng2020ansor} that use machine learning models such as XGBoost \cite{chen2016xgboost}. These models are characterized by simple structures and low computational costs. In recent years, researchers have begun experimenting with complex deep learning models, such as TIRAMISU \cite{baghdadi2021deep} and TLP using the dynamic computational flow LSTM model and  Transformer-based model, respectively.  The cost model does not take the tensor program directly, but the features extracted from tensor programs as input. These program features, such as the number of floating point add/multiply operators and the reuse distance of buffer access, etc, are often hand-selected by the compiler designer. To train these cost models, the compilers can use large offline datasets collected in advance, such as TenSet \cite{zheng2021tenset} providing a large-scale tensor program dataset on several hardware platforms, or small online datasets collected on the fly during the search, or both.  To address cross-platform online unawareness, some works \cite{zheng2021tenset, zhao2022moses, zhai2023tlp} also introduce many types of transfer learning, such as fine-tuning, distillation, training a local model to predict the gap between two domains, multi-task learning.

\subsection{Opportunities}
\textit{1) The exploration mechanism could be more efficient.} Table \ref{timecost} shows the costs of tuning a subset of DNNs on NVIDIA Jetson Orin with Ansor \cite{zheng2020ansor}, illustrating space exploration with learned cost model's time occupies nearly 40\%. The occupied time will increase when applying a more complex cost model. The exploration mechanism is expensive, due to extracting all explored tensor programs' features and feeding them to the learned cost model during each search round.

\begin{table}[h]
\footnotesize
\centering
\caption{Tuning costs (min) for Ansor with 2,000 trials on Orin, which means the time cost for space exploration, model training, and kernel hardware measurement, respectively.}
\label{timecost}
\begin{tabular}{c|ccc}
\toprule
\textbf{Ansor} & \textbf{R50} \cite{resnet} & \textbf{DeTR} \cite{detr}  & \textbf{I-V3} \cite{inceptionv3} \\
\midrule
Exploration & 35 & 30.31  & 41.8 \\
Training & 5.4 &  5.6  & 5.5 \\
Measurement & 44.4 & 50.61 & 49.4 \\

\bottomrule
\end{tabular}
\end{table}

In light of this, can we develop a simple draft model to roughly estimate performance, enabling an efficient "Draft-then-Verify" exploration mechanism?  Specifically, during the search phase, the draft model can quickly filter potential small-scale candidates, avoiding reliance on a complex and slow model. The learned cost model is then used only to verify and identify the optimal tensor program from these small-scale candidates rather than all explored candidates, greatly reducing inference overhead. Since tensor program performance aligns with the accelerator’s hierarchical parallel units,  we can design an empirical formula cost model as a draft model for initial exploration and draft small-scale candidates as input for the learned cost model’s output. This approach minimizes overhead and GPU usage.

\textit{2) Lack of easy-extraction and -training feature for deep learning-based cost model.} Recent works \cite{zhai2023tlp, zheng2021tenset, baghdadi2021deep} have demonstrated that deep learning-based cost models perform far better than other methods. However, they rely on complex expert feature engineering. TLP applies a Transformer-based model to capture the relation between temporal schedule primitives. The TLP model encodes schedule primitives using one-hot features, meaning that only a small portion of the feature vector (e.g., split factors) varies between different tensor programs—such as the differences are confined to only 1.387\% of the feature values in the case of a GEMM. This lack of diversity in the feature vectors makes it difficult to train the transformer model effectively on a small dataset. In our experiments (\S \ref{etetuning}), the TLP model fine-tuned on the target platform occasionally crashed, leading to tuning failures and the disappearance of the tuning curve.

Given this,  can we design a temporal feature that ensures the distinction between features of different tensor programs while accurately reflecting their behavior?  We think of the tensor program as a dataflow pipeline across hierarchical memory levels, encoding each data movement block into features that reflect corresponding computations, memory accesses, etc. Every value is related to its corresponding tensor program to ensure distinction and ease of training.   

\textit{3) Lack of effort-free adaptation strategy for deep learning-based cost model.} Some works \cite{zheng2021tenset, zhao2022moses} introduce distillation or transfer learning to address cross-platform online unawareness. However, these approaches require additional effort to complete the transfer tasks. Moses needs extra evaluation of the distillation process and TenSet additionally trains a local model to predict the gap, resulting in twice inference overhead. We introduce a momentum adaptation strategy compatible with any learned cost model, requiring no extra transfer overhead.

\begin{figure}[b]
  \centering
  \includegraphics[width=1\columnwidth]{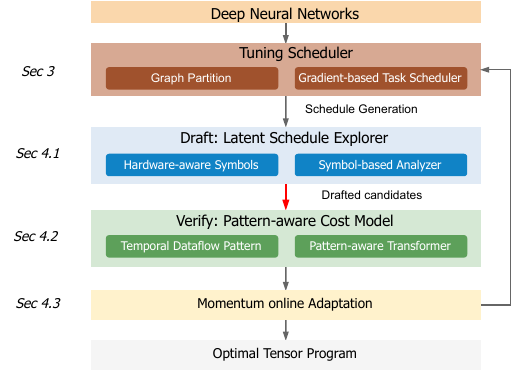}
  \caption{The system overview of Pruner, which contains latent schedule explorer and pattern-aware cost model.  Momentum online adaptation is only activated for MoA-Pruner in online cost model tuning scenarios.}
  \label{overview}
  \Description{}
\end{figure}

\section{System Design}
Figure \ref{overview} shows the system overview of Pruner, which is a "Draft-then-Verify"  exploration mechanism that accelerates the schedule search process. Pruner takes a DNN as input and converts it to partitioned small subgraphs using graph partitioning algorithm \cite{chen2018tvm}. Algorithm \ref{fgt} details the full graph tuning using MoA-Pruner.  Pruner uses Ansor’s gradient-based task scheduler \cite{zheng2020ansor} to tune these subgraphs over multiple rounds, with each round independently selecting one subgraph according to tuning record ($\mathcal{R}_{tune}$) (line 8). Pruner has two major components for "Draft" and "Verify", respectively: (1) latent schedule explorer (\textbf{LSE}) and (2) pattern-aware cost model (\textbf{PaCM}). One key challenge Pruner has to address is introducing a naive draft model to estimate performance roughly for rapid exploration. LSE (\S\ref{HAP}) designs the hardware-aware symbols and penalties to describe the utilization of hardware performance across different memory layers. LSE then introduces a symbol-based analyzer, an empirical formula cost model, and uses it to draft the small-scale candidates as an input ($\mathcal{S}_{spec}$) for the learned cost model during space exploration (line 9). To ensure some randomness, Pruner partially samples from the initial schedule space  (line 10).  The next challenge is to identify the best ($\mathcal{S}_{measured}$) from small-scale draft candidates ($\mathcal{S}_{draft}$).  PaCM (\S\ref{PaCM}) explores temporal dataflow feature, which is easy-extraction and -training for the transformer-based cost model, as a complement to naive statement-level features and designs the pattern-aware transformer, a multi-branch cost model, for accurate performance prediction (line 11). Then measure and update the tuning record (line 12). Finally, MoA-Pruner proposes a momentum online adaptation (\textbf{MoA}) strategy (\S\ref{lazyupdate}), introducing an online Siamese model to address the cross-platform online unawareness for online cost model tuning scenarios (line 13).

\begin{algorithm}[t]
    \footnotesize 
    \SetAlgoLined
    \LinesNumbered
    \SetAlgoNlRelativeSize{-1}
    \caption{Full Graph Tuning using MoA-Pruner}
    \label{fgt}
    % \tcp*[1]{1: $x$} 
    
    \KwIn{}
    \hspace{0.1cm} $P$: partitioned subgraph set \\
    \hspace{0.1cm} $d$: device abstraction \\
    \hspace{0.1cm} $nRounds$: number of rounds of tuning among all subgraphs \\
    \hspace{0.1cm} $C_{MoA}$: pre-trained cross-platform cost model \\
    
    \KwOut{$\mathcal{S}_{best}$}
    
    \SetKwFor{For}{for}{}{}  
    \SetKwFor{While}{while}{}{}
    \SetKwIF{If}{ElseIf}{Else}{if}{then}{else if}{else}{}  
    
    \SetKwFunction{sa}{$C_{SA}$}
    \SetKwFunction{SA}{$HaSchExplorer$}
    \SetKwFunction{fulltune}{$TuneFullGraph$}
    \SetKwProg{Fn}{Func}{\string :}{}% %定义python样式的函数格式

    \Fn{\fulltune{$P$, $d$, $C_{MoA}$}}{ 
        $\mathcal{R}_{tune} \leftarrow \emptyset, C_{PaT}\leftarrow C_{MoA}$\;
        \For{$i \gets 1$ \KwTo $nRounds$}{
            $p_0\leftarrow taskScheduler(P,\mathcal{R}_{tune})$\;
            % step 1
            $\mathcal{S}_{spec} \leftarrow LatentScheduleExplorer(p_0,d)$\;
            $\mathcal{S}_{draft} \leftarrow \mathcal{S}_{spec}\cup  RandomInitSch(p_0)$\;
            $\mathcal{S}_{measured} \leftarrow PaCostModel(C_{PaT}, \mathcal{S}_{draft})$\;
            $\mathcal{R}_{tune} \cup \{p_0:\mathcal{S}_{measured}\}$\;
            $C_{PaT}, C_{MoA} \leftarrow Update_{MoA}(C_{MoA}, \mathcal{R}_{tune})$\;
        }
        $\mathcal{S}_{best} \leftarrow BestSchedules(\mathcal{R}_{tune})$\;
        \Return{$\mathcal{S}_{best}$}\;
    }
\end{algorithm}

\section{Pruner}

\subsection{Draft: Latent Schedule Explorer}
\label{HAP}

During the search process, our primary objective is to employ an efficient draft model for the rapid schedule space exploration. We observe that tensor program performance on hardware aligns with the hierarchical parallel units of the accelerator. To design an efficient empirical formula cost model (draft model), we focus on quantitatively analyzing the impact of these assignments of schedule primitives on performance. Pruner introduces Latent Schedule Explorer (LSE). LSE  formulates the search process as a hardware fitness optimization problem, relying on the draft model rather than the learned cost model.

\begin{figure*}[t]
\centering
\centerline{\includegraphics[width=0.92\linewidth]{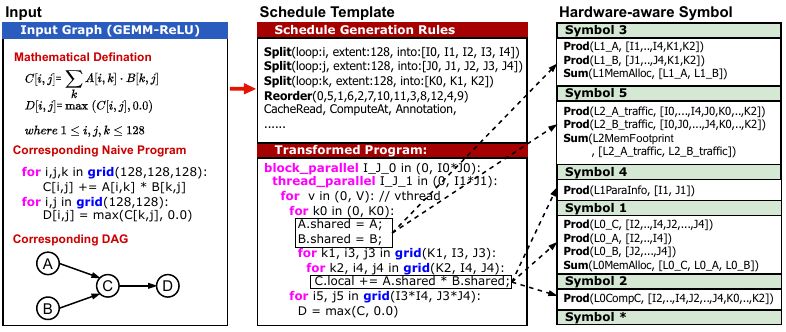}}
\caption{An illustrative example of the hardware-aware symbols extraction process for a GEMM-ReLU graph. Some schedule primitives in the schedules generation rules and hardware symbols for some statements are omitted for brevity. Prod or Sum means that the variable equals the product or sum of an array of variables.} 
\label{sampleforsymbol}
\Description{}
\end{figure*}

% √
Algorithm \ref{se} details how Pruner constructs a small-scale schedule candidate set ($\mathcal{S}_{spec}$) during LSE. The inputs to the explorer include the subgraph $p_0$, the corresponding computation graph (i.e., DAG),  and the device abstraction. First, Pruner forms the schedule space ($\theta_x$) using schedule generation rules \cite{zheng2020ansor} and randomly samples a set of initial schedules ($\mathcal{S}_x$) on lines 13 and 14. Lines 16 to 21 introduce a draft model (Symbol-based Analyzer), which runs for nSteps to optimize the hardware fitness scores. Each step estimates the performance of the current value $\mathcal{S}_{x}$ using hardware-aware symbols and Symbol-based Analyzer (lines 5 to 11).  Then Pruner uses a genetic algorithm (GA) to update $\mathcal{S}_{x}$ in each step (line 21). GA  explores tiling-factor transformations for for-loops, with a fitness function using the draft model’s estimated performance to guide the mutation.  Finally, Pruner outputs the $\mathcal{S}_{spec}$ with the highest hardware fitness scores.

\begin{algorithm}[h]
    \footnotesize 
    \SetAlgoLined
    \LinesNumbered
    \SetAlgoNlRelativeSize{-1}
    \caption{Latent Schedule Explorer}
    \label{se}
    % \tcp*[1]{1: $x$} 
    
    \KwIn{}
    \hspace{0.1cm} $p_0$: subgraph to be optimized \\
    \hspace{0.1cm} $d$: device abstraction \\
    \hspace{0.1cm} $nSteps$: number of steps to run GA \\
    \hspace{0.1cm} $perfCost$: Symbol-based Analyzer \\
    
    \KwOut{$\mathcal{S}_{spec},\mathcal{\theta}_x$}
    
    \SetKwFor{For}{for}{}{}  
    \SetKwFor{While}{while}{}{}
    \SetKwIF{If}{ElseIf}{Else}{if}{then}{else if}{else}{}  
    
    \SetKwFunction{sa}{$C_{SA}$}
    \SetKwFunction{SA}{$LatentScheduleExplorer$}
    \SetKwProg{Fn}{Func}{\string :}{}% %定义python样式的函数格式

    \Fn{\sa{$sch$, d}}{ 
            $symbols\leftarrow \emptyset$, $cost \leftarrow$ 0\;
            \For{$stmt \in \  \tau$($p_0$, $sch$).bufferStmt}{
                \For{$symbol \in \  hardware\_aware\_symbols$}{
                    $symbols$.append($symbol$($stmt$))\;
               }
               $penalties,\ utilization \leftarrow hardware\_aware\_penalty(symbols, d)$\; 
               % $utilization \leftarrow hardware\_utilize(penalties,d)$\; 
               
               $cost\leftarrow$ $cost+ perfCost(utilization, symbols$)\;
           }
        \Return{$cost$}\;
    }
    
    \Fn{\SA{$p_0$, $d$}}{ 
        $\mathcal{\theta}_x \leftarrow GenerateSketch(p_0$)\;
        $\mathcal{S}_{x} \leftarrow   RandomInitSch(\theta_x$)\;
        $\mathcal{S}_{spec} \leftarrow \emptyset$\;
        \For{$i \gets 1$ \KwTo $nSteps$}{
            $costs\leftarrow \emptyset$\;
            \For{$sch \in \   \mathcal{S}_{x}$}{
              $costs \leftarrow costs \cup (sch, C_{SA}(sch,d))$\;
            }
            $\mathcal{S}_{spec} \leftarrow PriorFilter( \mathcal{S}_{x}\cup \mathcal{S}_{spec},costs)$\;
            $\mathcal{S}_{x} \leftarrow SchMutation(\mathcal{S}_{x}, \theta_x, costs)$\;
        }
        \Return{$\mathcal{S}_{spec},\mathcal{\theta}_x$}\;
    }
\end{algorithm}

\begin{table}[b]
\footnotesize
\centering
\caption{Hardware-aware symbols and related memory level.}
\label{hasymbol}
\begin{tabular}{cl}
\toprule
\textbf{Mem} & \textbf{Symbol}  \\
\midrule

L0 & S1-L0MemAlloc, S2-L0CompCount \\
L1 & S3-L1MemAlloc, S4-L1ParaInfo  \\
\multirow{2}{*}{L2} & S5-L2MemFootprint, S6-L2ParaInfo \\
                    & S7-L2TransDim, S8-L2CompCount \\

\bottomrule
\end{tabular}
\end{table}

\noindent\textbf{Hardware-aware Symbols.} Based on the common characteristics of generated schedule primitives, Pruner extracts hardware-aware symbols to describe the program's behaviors in hierarchical memory. Table \ref{hasymbol} presents hardware-aware symbols generated based on schedule primitives. Concretely, Symbols 1 and 3 count the allocation of L0 and L1 level storage, respectively.  Symbol 2 describes the total amount of computation at the L0 level. Symbols 4 and 6 describe the parallel characteristics of the program. Symbol 5 counts the memory footprint of the lowest-level storage. Symbols 7 and 8 describe the innermost dimension length and total amount of computation at the L2 level.

As a specific example, Figure \ref{sampleforsymbol} illustrates the hardware-aware symbol extraction process for a GEMM-ReLU fused operator during GPU compilation. This process involves three main steps: the input graph, schedule template generation, and hardware-aware symbol extraction. The input graph consists of the original program and its associated DAG. Next, a corresponding schedule template is generated by applying rules to the stage nodes of the DAG in reverse topological order, similar to Ansor. Finally, Pruner traverses all statements to extract hardware-aware symbols.

\noindent\textbf{Hardware-aware Penalty.} Pruner converts these symbols into six penalty terms, $\boldsymbol{\mathcal{P}_{li,*}}$, which reflect how the behavior of tensor programs at hierarchical memory levels impacts the utilization of the hardware's theoretical peak performance. Here, $li$ and $*$ represent the memory level and the penalty type, respectively, with the latter including computation (c) and memory (m). For symbols involving memory capacity, we use piecewise functions to quantify their impact on performance. For example, there is an upper limit $m_{l0}$ at the L0 level; if the allocation $S1$ exceeds $m_{l0}$, it will incur data transmission overhead. Therefore, we define $\mathcal{P}_{l0,m} \coloneqq \min{\left(\frac{m_{l0}}{S1}, 1\right)}$.

\textit{L0 level.} In addition to the $\mathcal{P}_{l0,m}$ we described earlier, we also define a  compute-to-memory penalty at this level  as $\mathcal{P}_{l0,c} \coloneqq1+\frac{S2}{S1}$. The bigger it means less memory allocation and higher computing efficiency.

\textit{L1 level.} Like $\mathcal{P}_{l0,m}$, Pruner defines $\mathcal{P}_{l1,m} \coloneqq \min{(\frac{m_{l1}}{S3},1)}$, where $m_{l1}$ represents the L1 memory allocation limit.   Issues such as computation scheduling (e.g., warp scheduling in GPUs) are involved. The degree of alignment between the program scheduling and the hardware parallel execution unit determines the utilization of hardware performance. Pruner defines $\mathcal{P}_{l1,c} \coloneqq sch_{l1} / (\lceil {\frac{sch_{l1}}{pu_{l1}} \rceil \cdot pu_{l1}})$  to describe the utilization of parallel resources by scheduling at L1 level,  where $sch_{l1}=\lceil \frac{S4}{n_{l1}} \rceil$ refers to the number of scheduling blocks,  $pu_{l1}$ and $n_{l1}$ refer to the number of L1 blocks that can be activated simultaneously and the scheduling size within a block (e.g., warp size in GPUs) at the L1 level, respectively. As a supplement, Pruner also describes scheduling waste issues as $\alpha_{l1} \coloneqq S4 / (sch_{l1}\cdot n_{l1})$.

\textit{L2 level.} Similar to penalty term $\mathcal{P}_{l1,c} $, Pruner defines $\mathcal{P}_{l2,c} $ to describe the utilization of parallel units at the lowest level (e.g., SMs in GPUs) during the execution of tensor programs. The definition as $\mathcal{P}_{l2,c} \coloneqq S6 / (\lceil {\frac{S6}{pu_{l2}} \rceil \cdot pu_{l2}})$,  where   $pu_{l2}$   refers to L2 blocks that can be scheduled simultaneously  (e.g., SMs in GPUs) at L2 level. We consider the access transactions to lowest memory level and define the $\mathcal{P}_{l2,m}$ term as $\mathcal{P}_{l2,m}\coloneqq S7 / (\lceil {\frac{S7}{n_{l2}}} \rceil \times n_{l2})$, where $n_{l2}$  represents memory transaction length at L2 level.

\noindent\textbf{Symbol-based Analyzer.}   During the Latent Schedule Explorer, Pruner needs a draft model to evaluate the performance of the schedule to rapidly guide the optimization of the  $\mathcal{S}_{spec}$'s hardware fitness. Inspired by static code analysis, Pruner proposes an empirical formula named Symbol-based Analyzer (\textbf{SA}) instead of using a deep learning-based cost model during exploration.

With the hardware-aware penalty terms $\mathcal{P}_{li,*}$, SA can easily derive the performance of a schedule. First, SA extracts the hardware utilization for each innermost statement in tensor programs. For computation-related statements, the utilization ($U_{p}$) of the hardware's theoretical peak performance ($T_{p}$) can be estimated across different level penalty terms $\mathcal{P}_{li,c}$, deriving as $U_{p} = T_{p} \cdot \prod_{li}{\mathcal{P}_{li,c}}$. For example, given a schedule with six scheduling blocks and hardware with four parallel units at $li$ level, the utilized performance can be estimated as $0.75\cdot T_{P}$. To simplify the assessment ($U_{m}$) of memory bandwidth ($T_{m}$), we only consider the L2 storage level, which has the highest latency, and incorporate the lower-level penalties as $U_{m} = T_{m} \cdot \prod_{li}{\mathcal{P}_{li,m}}$. Second, SA can estimate the computation and memory access latency of each innermost statement according to the Eq. \ref{hdlatency}, obtaining the total latency $L_{total}$ of each tensor program, where $S8$ and $S5$ (see Hardware-aware Symbols) denote the number of float operators and the actual memory access of $i$-th statement.
\begin{equation}
\label{hdlatency}
% \small
\centering
    \begin{aligned}
    & L^i_c=\frac{S8}{U_p},\ \  L^i_m=\frac{S5}{U_{m}}, \ \   L_{total}=\sum_{i}{(L_{c}^{i} + L_{m}^i)}
    \end{aligned}
\end{equation}

%-----
\subsection{Verify: Pattern-aware Cost Model}
\label{PaCM}

\begin{figure}[b]
  \centering
  \includegraphics[width=\columnwidth]{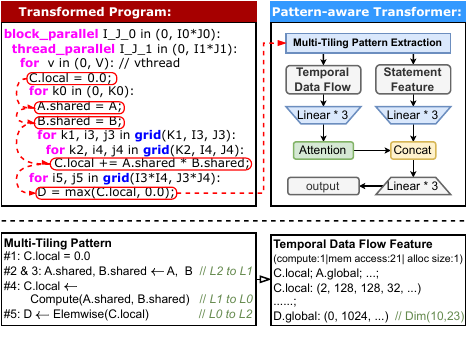}
  \caption{(top) The pipeline of Pattern-aware Cost Model; (bottom) Extraction of temporal dataflow feature.}
  \label{modelpipeline}
  \Description{}
\end{figure}

After generating the $\mathcal{S}_{spec}$ by Latent Schedule Explorer (\S \ref{HAP}), the next goal of Pruner is to build an efficient and accurate cost model to identify the optimal tensor program. The current mainstream approach is to use a trained machine learning or deep learning model as the cost model. Apart from the different models used (such as XGBoost \cite{chen2016xgboost} and MLP), the primary distinction among these methods lies in the definition and extraction of program features. AutoTVM \cite{chen2018learning} aims to extract features from each loop variable, while Ansor \cite{zheng2020ansor} and TenSetMLP \cite{zheng2021tenset} leverage the features of the innermost statements. However, feature designs specified to the single variable or statement fail to adequately characterize the behaviors of tensor programs, leading to limited performance. In addition to feature design, TLP \cite{zhai2023tlp} uses a Transformer-based model to predict the performance of schedule primitives. However, training this model requires a large external dataset and significant computational resources for inference. We observe an insight that the multi-tiling pattern recurs in different tensor programs, representing the dataflow process between hierarchical memory. Pruner designs a Pattern-aware Cost Model (\textbf{PaCM}), including the temporal dataflow feature recognition across hierarchical memory and a resource-efficient Transformer model.  Therefore, we attempt to define the temporal dataflow features of tensor programs as complementary to the naive statement features for further enhancing the prediction accuracy of the cost model. Specifically, we extract critical dataflow and their features from the low-level intermediate representation (IR) of tensor programs using the multi-tiling pattern, then design a multi-branch Pattern-aware Transformer to learn the modeling these serialized dataflow features and the mapping from features to performance, as shown in Figure \ref{modelpipeline}.

\noindent\textbf{Feature Representation.} Most tensor programs consist of nested loops with assignment statements, aiming to apply the multi-level tiling rules to improve data reuse and computational performance. As illustrated in Figure \ref{modelpipeline} bottom, we first abstract a multi-tiling pattern covering multi-level buffers across different memory levels (e.g., register, shared, and global memory in GPUs), to extract the data block movement process with temporal information from tensor programs.  Then, Pruner treats statements as temporal dataflow blocks across hierarchical-level memory and encodes behaviors involving different buffers separately into a 23-dimensional embedding vector to define the dataflow features of each tensor program. For example in Figure \ref{modelpipeline} bottom, the statement ($C.loacl\leftarrow Compute(A.shared, B.shared)$)  covers three data block movements. Each data block’s characteristics include memory accesses(e.g., data reuse, access stride, access type),  allocation sizes, flow direction (e.g., L2->L1), and computation density, etc. Those values will be calculated based on the buffer information and for-loops bound to them. Given that the multi-tiling pattern in different programs shares the same data movement process except for values, the resulting structured features facilitate the convergence of subsequent cost models.  Element-wise operators don’t require tiling to improve data reuse, for which tiling templates cannot capture dataflow characteristics. Besides, Pure element-wise operators accounting for less than 3\% (based on the TenSet), are typically fused into tiled operators in DNNs. Therefore we apply a zero-padding simply for element-wise op’s features,  requiring no additional computational overhead.  Finally, combining the statement-level features provided by Ansor, we construct a hybrid feature to describe the behavior of the tensor program.

\noindent\textbf{Pattern-aware Transformer.} To fully exploit the rich semantics of the hybrid feature, we propose a multi-branch Pattern-aware Transformer, as illustrated in Figure \ref{modelpipeline}. For statement-level features, we encode them using multiple linear layers, followed by summation to obtain a high-dimension vector. As for temporal dataflow features, considering the inherent strong contextual correlation and temporal information, we employ a self-attention mechanism \cite{vaswani2017attention} to model context dependencies. Finally, PaCM outputs normalized predictions through a concatenation operator and multiple linear layers. To train the model, we use the normalized latency and LambdaRank loss \cite{cao2007learning,wang2018lambdaloss} as the ground truth and optimization objective.

\begin{figure}[b]
\begin{center}
\centerline{\includegraphics[width=1\columnwidth]{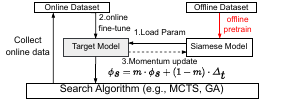}}
\caption{Overview of MoA, where $\mathcal{\phi}_{s}$ and ${\Delta}_{t}$ refer to the parameters of the Siamese and gradient of target model, the momentum $m=0.99$. The red arrow means offline pre-train.}
\label{siameseupdate}
\Description{}
\end{center}
\end{figure}

\subsection{MoA-Pruner}
\label{lazyupdate} 
\textbf{Momentum online Adaptation.} Although deep learning-based cost models can achieve satisfactory performance, their high dependency on training data poses significant challenges for cross-platform online unawareness. A cost model trained on one hardware platform usually performs poorly on another and cannot be applied to another online. When applying pre-trained models to new hardware platforms, most learning-based cost models typically employ strategies including transfer learning \cite{zheng2021tenset} and knowledge distillation \cite{zhao2022moses}. However, these strategies often yield limited effects due to the extra transfer overhead. For online finetuning, the limited and biased data collected during the early stages of online training can disrupt model training. Current research on cost models has also not effectively addressed the challenge of model transfer in online training scenarios. To address this issue, we propose a momentum online adaptation  (\textbf{MoA}) strategy based on a Siamese network that has the same model architecture and similar parameters, as illustrated in Figure \ref{siameseupdate}. During the online cost model update of each tuning round, we treat the cross-platform pre-trained cost model as a Siamese network and use its weights to initialize the target model, then fine-tune it through collected online data. Notably, we finally use a momentum update strategy according to the gradients of the target model to adjust the weights of the Siamese network, similar to MoCo \cite{he2020momentum, chen2020improved, chen2021exploring}, without the need for the Siamese network's forward and backward.  Siamese models offer high-quality initial weights, simplifying training. Through momentum gradient updates, these weights of Siamese models adapt to the target platform, further optimizing future training. This bidirectional feedback mitigates this disruption and reduces the difficulty of model training in online tuning scenarios.  Experimental results demonstrate that this method can ensure the stability of the training process and achieve better convergence compared to existing methods with the same scale of online collected fine-tuning data.

\section{Experimental Settings}
\textbf{DNN workloads.} Pruner is evaluated on 14 DNN models for inference scenarios, covering various computer vision and language models. Table \ref{dnn_a} describes their input shape and precision and table \ref{dnn_b} lists the details of the language models. These networks contain a large part of operators commonly seen in DNNs. In addition to full-precision (F), we also select six LLMs for half-precision (H) optimization on TensorCore.

\begin{table}[b]
    \centering
    \caption{Evaluated DNN models in Pruner, with shapes and optimization precisions.}
    \label{dnn_a}
    \footnotesize
    \resizebox{\linewidth}{!}{
        \begin{tabular}{cc|cc}
            \toprule
            $\rm \textbf{CNNs}$ & $\rm \textbf{Shape \& Precision}$ & $\rm \textbf{Transformers}$ & $\rm \textbf{Shape \& Precision}$ \\
            \midrule
            ResNet\cite{resnet} & (1, 3, 224, 224) \& (F) & Bert-B/T\cite{bert} & (1 \& 4, 128) \& (F, H)\\
            WideResNet\cite{zagoruyko2016wide} & (1, 3, 224, 224) \& (F) & GPT-2\cite{radford2019language} & (1 \& 4, 128) \& (F, H) \\
            Inception-V3\cite{inceptionv3} & (1, 3, 299, 299) \& (F) & Llama\cite{touvron2023llama} & (1 \& 4, 128) \& (F, H) \\
            Densenet-121\cite{densenet} & (1, 3, 224, 224) \& (F) & OPT\cite{zhang2022opt} & (1 \& 4, 128) \& (H) \\
            Mobilenet-V2\cite{mobilenetv2} & (1, 3, 224, 224) \& (F) & Mistral\cite{jiang2023mistral} & (1 \& 4, 128) \& (H)\\
            DCGAN\cite{radford2015unsupervised} & (1, 100) \& (F) & ViT\cite{vit} & (1, 3, 256, 256) \& (F)  \\
            Deeplab-V3\cite{deeplabv3} & (1, 3, 224, 224) \& (F) & DeTR\cite{resnet} & (1, 3, 256, 256) \& (F) \\
            \bottomrule
        \end{tabular}
    }
\end{table}

\begin{table}[b]
    \centering
    \caption{Details for Transformer-based language models.}
    \label{dnn_b}
    \footnotesize
    \begin{tabular}{l|cccc}
        \toprule
        \textbf{Model} & \textbf{layers} & \textbf{heads} & \textbf{hidden} & \textbf{intermediate} \\ 
        \midrule
        Bert-Tiny      & 6                    & 8                               & 512                  & 2048                        \\
        Bert-Base      & 12                   & 12                              & 768                  & 3072                        \\ 
        GPT-2          & 12                   & 12                              & 768                  & 3072                        \\ 
        OPT-1.3b       & 24                   & 32                              & 2048                 & 8192                        \\ 
        Llama          & 12                   & 12                              & 768                  & 3072                        \\ 
        Mistral-7b     & 32                   & 32                              & 4096                 & 14336                       \\ 
        \bottomrule
    \end{tabular}
\end{table}

\begin{figure*}[t]
\begin{center}
\centerline{\includegraphics[width=1.0\linewidth]{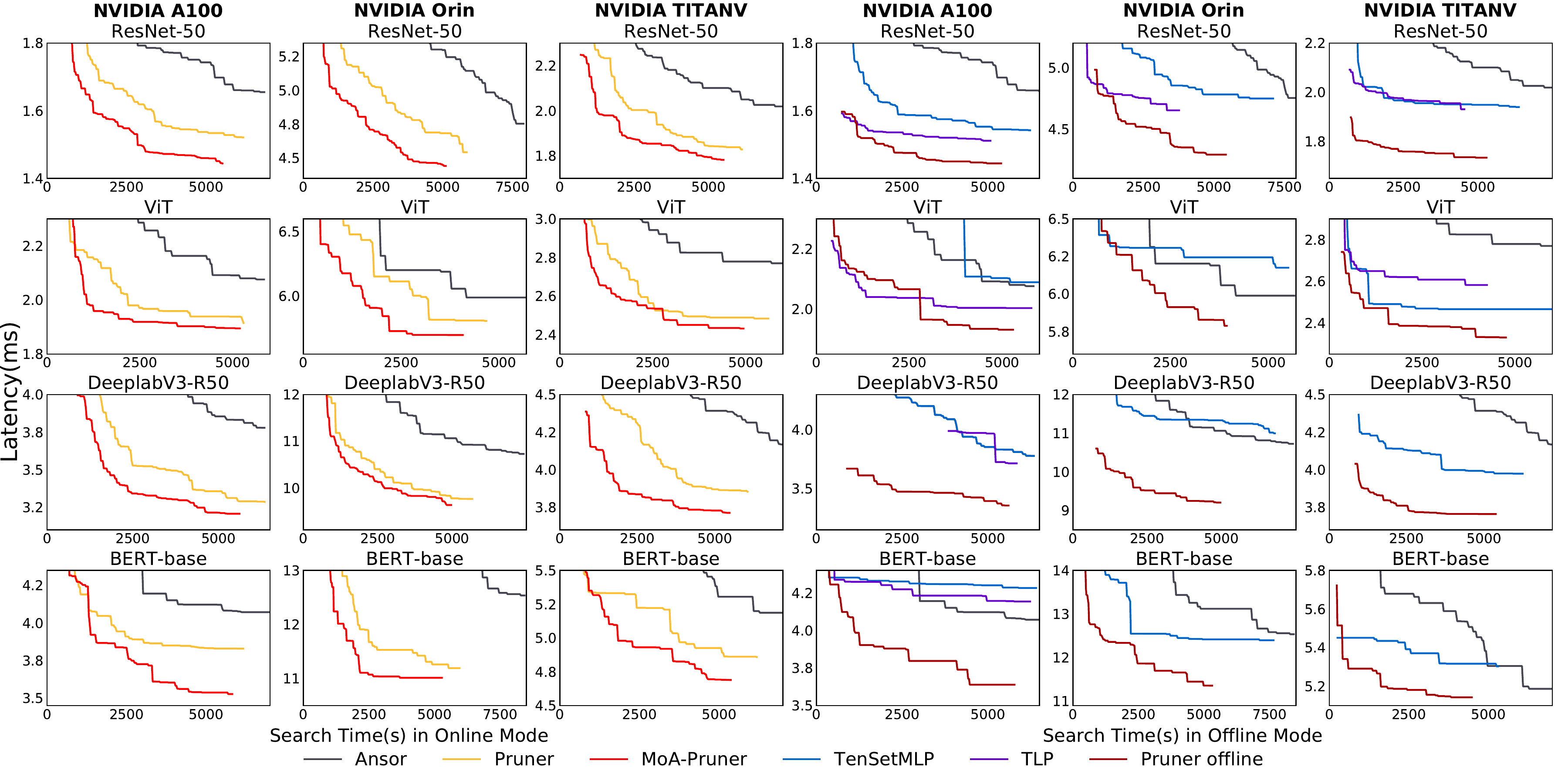}}
\caption{Workload tuning curves in online and offline cost model tuning mode.  Each row takes the same workload, and each column takes the same platform.}
\label{end-to-end-tuning-curve}
\Description{}
\end{center}
\end{figure*}

\noindent\textbf{Platforms and baselines.} We evaluate Pruner on three representative NVIDIA GPU platforms: A100, Titan V, and  Jetson Orin-AGX. The A100 and Titan V represent devices used in servers,  while the Jetson Orin-AGX represents devices commonly found in edge computing environments.  We compare Pruner with three off-the-shelf inference frameworks: PyTorch 2.2,  Triton \cite{tillet2019triton}, and Torch-TensorRT \cite{TRtorch}, as well as eight SOTA tensor compilers: Ansor \cite{zheng2020ansor}, TenSetMLP \cite{zheng2021tenset}, TLP \cite{zhai2023tlp}, MetaSchedule \cite{shao2022tensor}, Felix \cite{zhao2024felix}, Adatune \cite{li2020adatune}, Roller \cite{zhu2022roller} and TLM \cite{zhai2024enabling}. We utilize the TorchInductor and Torch-TensorRT backends in PyTorch by using the command torch.compile(backend=..., mode=...). For the TorchInductor backend, we use different modes such as "max-autotune" (for the Triton kernel) and "reduce-overhead".

\noindent\textbf{Tuning settings.}  We discuss the effectiveness of Pruner in both offline and online cost model tuning scenarios. For the offline tuning mode, Pruner, TLP, and TenSetMLP are all pre-trained on the TenSet GPU dataset and fine-tuned on the target platform dataset. For each platform, we built datasets containing approximately 500,000 tensor programs. For the online mode, all search-based baselines update the cost model through online data collection.  The Siamese model used by MoA is pre-trained on the TenSet GPUs K80-6M dataset. During the tuning process, our experiment setting for search-based compilers is similar to TLP: we set the maximum number of rounds to 200, selecting ten tensor programs for evaluation in each round, totaling 2,000 trials. We also compared Pruner's performance under 2,000 trials with Ansor's tuning performance under more trials. Based on the subsequent experiments, we set the size of drafted candidates ($\mathcal{S}_{spec}$ generated by LSE)  to 512.

\section{Evaluation}

\subsection{End-to-End Workload Benchmark}
\label{etetuning}
We evaluate the tuning performance of Pruner on end-to-end workloads and compare it with search-based DLCs and DNN frameworks in terms of tuning efficiency and effectiveness.

\noindent\textbf{Insight on Pruner’s fast convergence.} The tuning efficiency and quality of the end-to-end workload can directly reflect the overall performance of search-based DLCs. We evaluate the Pruner on 10 different workloads and compare it with Ansor \cite{zheng2020ansor}, TenSetMLP \cite{zheng2021tenset}, and TLP \cite{zhai2023tlp}. Considering that the tuning process of existing methods includes both offline and online modes, we discuss the effectiveness of Pruner in these two cases, respectively. This section pertains to the experimental setting we previously discussed. In this part, all methods tune the DNNs with a total of 2,000 trials.

Figure \ref{end-to-end-tuning-curve} illustrates the tuning curves of different methods on different GPUs under online and offline modes for a subset of DNNs. In some cases, the tuning curve of TLP disappears because it fails to search for an available solution after fine-tuning. We observe that in both tuning scenarios, Pruner consistently searches better schedules faster than other baselines, as evidenced by its quicker convergence to lower values on the tuning curve.  Regarding tuning time, Pruner completes the tuning task earlier than others given the same tuning trials. This advantage is due to Pruner's LSE not relying on the learned cost model and reducing the time overhead on cost model inference. In terms of tuning performance, Pruner exhibits a significant gap compared to other baselines, and this gap emerges early in the tuning process, especially in online cost model tuning scenarios. The main reason is that  LSE facilitates the initial screening of schedule space and enables the identification of better schedule sets during the exploration process even using a naive draft model. During the entire tuning process, Pruner consistently achieves steady improvements, thanks to LSE's rapid exploration of the schedule search space and PaCM's more accurate performance modeling. MoA further enhances PaCM training in online scenarios, leading to faster convergence of the tuning curve.

\begin{figure}[b]
\begin{center}
\centerline{\includegraphics[width=1\columnwidth]{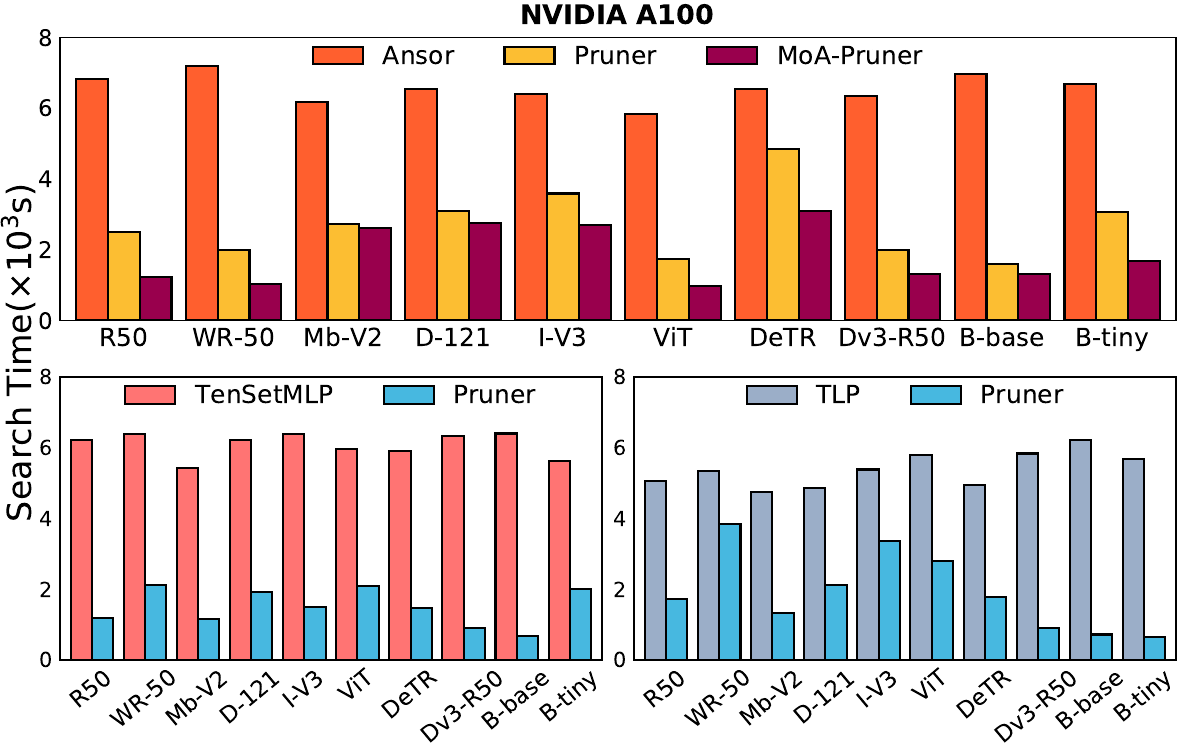}}
\caption{Search time comparison, measured by how much time Pruner takes to reach the best Ansor/TenSetMLP/TLP’s entire search performance on  NVIDIA A100, respectively.}
\label{wopretrain_compile_time_a100}
\Description{}
\end{center}
\end{figure}

We recorded the search time required for Pruner to achieve the same tuning performance as other baselines on both offline and online modes. In online scenarios,  Pruner can obtain average speedups of $2.7\times$, $2.5\times$, and $2.59\times$ on the three platforms, respectively. More importantly, when we use the MoA strategy to introduce the cross-platform pre-training model, the average speedups of MoA-Pruner can be increased to $4.18\times$, $4.77\times$, and $5.51\times$, respectively. In offline scenarios, compared with TenSetMLP, Pruner can achieve average speedups of $4.67\times$, $4.53\times$, and $5.06\times$ on the three platforms, respectively. Due to TLP's inability to search in some workloads, for the sake of fairness, we only compare Pruner against TLP on A100, achieving an average speedup of $4.05\times$.   An example, Figure \ref{wopretrain_compile_time_a100} shows the search time comparison, measured by how much time Pruner takes to reach the best Ansor/TenSetMLP/TLP’s entire search performance on A100, respectively, proving the effectiveness and superiority of proposed methods.

\textbf{Pruner's stable performance improvements.} In assessing the degree of performance enhancement realized by Pruner, we conducted further comparisons on a subset of DNNs with Ansor, which has more tuning trials, on the NVIDIA A100. We comprehensively compared their tuning performance and the time required for tuning. Table \ref{moretrials} illustrates that even with 2,000 trials, Pruner usually outperforms Ansor with more trials in terms of search time and tuning performance, in line with our earlier observations. As a complement, we also conducted a comparative experiment with the representative method TenSetMLP to demonstrate MoA's effectiveness. Table \ref{moretrials} shows our MoA has advantages in both search time and tuning quality.

\begin{table}[t]
\footnotesize
\centering
\caption{The tuning performance (ms) and compilation cost (min) of MoA-Pruner (with 2k trials) compared to Ansor (with more trials) and TenSet's transfer strategy (with 2k trials) on NVIDIA A100.}
\label{moretrials}
\resizebox{\linewidth}{!}{
\begin{tabular}{c|ccc|cc|cc}
\toprule
 \multirow{2}{*}{Models} & \multicolumn{3}{c}{Ansor} &  \multicolumn{2}{c}{TenSet's transfer} & \multicolumn{2}{c}{MoA-Pruner} \\
    \cmidrule{2-8}
  & trials & perf & cost & perf & cost & perf & cost\\
\midrule
 ResNet50 & 10k & 1.458 & 743 & 1.817 & 131 & \textbf{1.444}  &  \textbf{91}\\
 Inception-v3 & 10k & \textbf{2.694} & 739 & 3.493 & 128 & 2.739  & 93 \\
 Bert-Base & 6k & 3.872 & 462 & 5.287 & 136 & \textbf{3.527}  & \textbf{98}\\
Bert-Tiny & 6k & 1.413 & 441 & 1.573 & 122&  \textbf{1.27}  & \textbf{84} \\
\bottomrule
\end{tabular}
}
\end{table}

\begin{figure}[b]
\begin{center}
\centerline{\includegraphics[width=1\columnwidth]{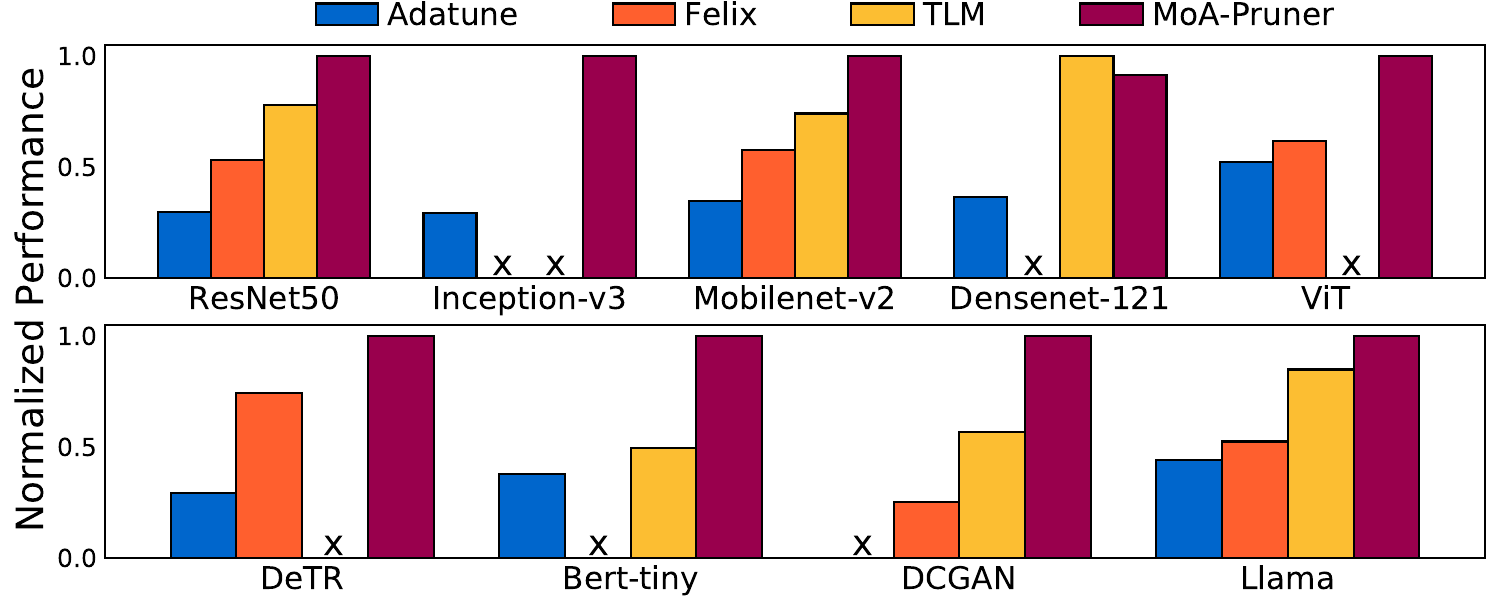}}
\caption{Normalized inference performance comparison with more tensor compilers on NVIDIA A100.}
\label{comparewithbaseline}
\Description{}
\end{center}
\end{figure}

\begin{table}[b]
\footnotesize
\centering
\caption{Workload inference latency (ms) comparison with Roller on TITAN V.}
\label{roller}
\resizebox{\linewidth}{!}{
\begin{tabular}{c|c|cccc}
\toprule
 $\rm \textbf{Models}$ & Input Shape & Pytorch & Roller & Ansor & MoA-Pruner \\
\midrule
\multirow{2}{*}{ResNet50} & (1, 3, 224, 224) & 7.01 & 4.72 & 2.245  & \textbf{1.886} \\
 & (128, 3, 224, 224) & 126.02 & 136.15 & 115.52  & \textbf{101.01} \\
Bert-Large & (1, 128) & 26.5 & 18.04 & 21.658 & \textbf{17.533} \\
\bottomrule
\end{tabular}
}
\end{table}

\textbf{Pruner vs. more tensor compilers.} We compared Pruner with three search-based tensor program compilers: Adatune \cite{li2020adatune}, Felix \cite{zhao2024felix}, and TLM \cite{zhai2024enabling}. Figure \ref{comparewithbaseline} presents the tuning performance, expressed as the normalized speedup relative to the best inference time,  achieved by Pruner and these methods on the A100 GPU. In some cases, those methods encounter challenges in tuning effectiveness, which leads to failure (denoted as $X$ in the figure). For instance, Adatune lacks support for ConvTranspose2d, Felix struggles with operators of special cases and irregular shapes due to its unique feature extraction mechanism, and TLM, as a language model, only supports the subgraphs collected in its pre-training dataset. When we applied it to a model that didn't appear in the training phase, it failed to tune. For the sake of fairness, we only compare Pruner with others in their successful tuning cases, and the average speedup over other methods achieved by MoA-Pruner is $1.37\times$, $1.85\times$, and $2.77\times$ for TLM, Felix, and Adatune, respectively. Pruner shows stable search performance in end-to-end tuning scenarios against other search-based tensor compilers, thanks to its versatile feature design, whether in LSE or PaCM, Pruner is not limited to specific operators or shapes, ensuring consistent and stable search performance for diverse models.

We also compared Pruner with Roller \cite{zhu2022roller} by utilizing the Docker image (which only supports V100 or older GPUs) on TITAN V for our experiments.  As a rule-based tensor compiler, Roller runs fast but easily misses optimal solutions (confirmed in the TLM paper\cite{zhai2024enabling} and our experiments).  We set the tuning trials to 50 per subgraph and report the tuning latency in Table \ref{roller}. Results show Pruner has the lowest tuning latency and maintains stable search quality and efficiency, due to LSE's quick exploration of the space without prematurely discarding solutions outside the rules.

\begin{figure}[b]
\begin{center}
\centerline{\includegraphics[width=1\columnwidth]{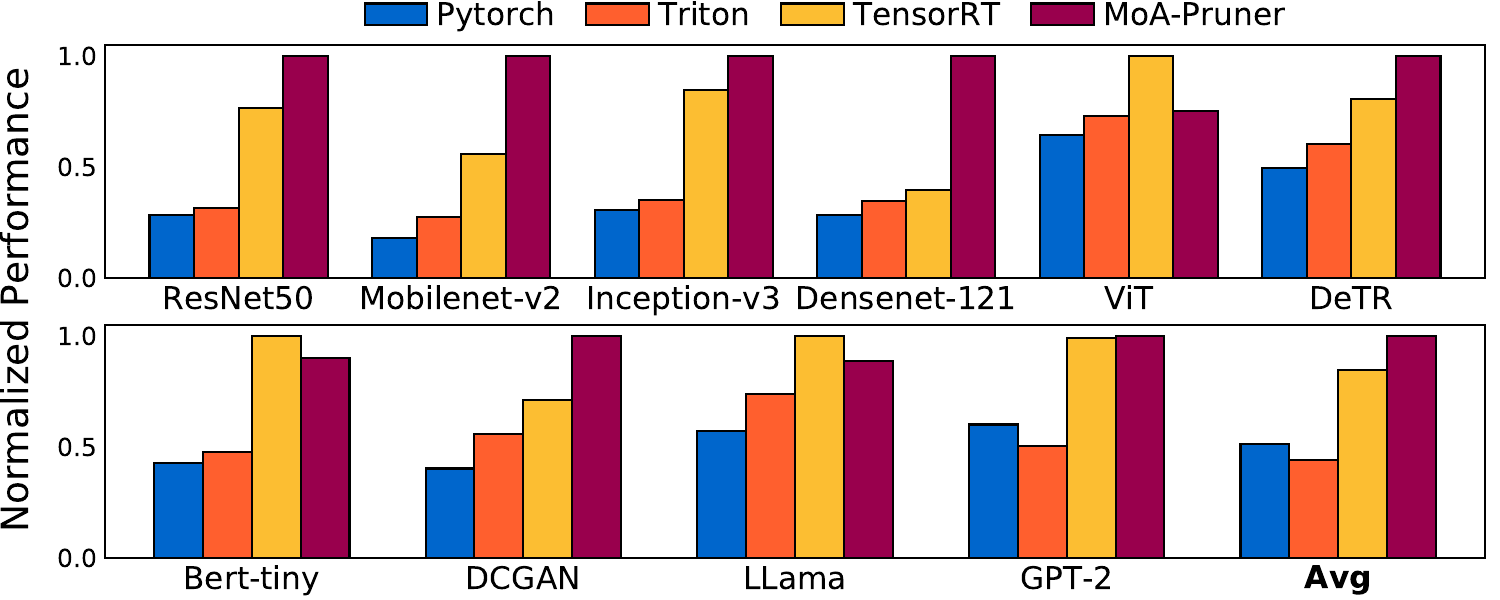}}
\caption{Normalized inference performance comparison with off-the-shelf inference frameworks on NVIDIA A100.}
\label{comparewithFramwork}
\Description{}
\end{center}
\end{figure}

\textbf{Pruner vs. off-the-shelf inference frameworks.} Figure \ref{comparewithFramwork} presents the normalized inference performance of 10 DNNs on A100 when using Pruner and inference frameworks. The average speedup over DNN frameworks achieved by Pruner is $1.95\times$, $2.27\times$, and $1.21\times$ for PyTorch2.2, Triton, and TensorRT, respectively. We found that the speedup achieved by Pruner depends on the type of operators in the DNNs. For instance,  TensorRT outperforms Pruner in some cases.  One reason we identify is that the speedup achieved by TensorRT relies on deep optimization of handwrite kernel libraries, as well as the utilization of hardware computing units. When the parallel axis dimension is small (e.g., patch blocks in ViT $n$ is 64) but the reduction axis dimension in matrix multiplication is large (e.g., 2048), kernel libraries leverage techniques like sliced1x4 or splitK to optimize for the reduction axis. In contrast, TVM prioritizes generality by employing simple multi-level tiling rules to construct the search space, which may limit performance in specialized scenarios. For instance, in a linear operator of ViT with input shape (1, 65, 2048) and weight shape (2048, 1024), Pruner achieves 41.3$\mu$s, slightly underperforming cuBLAS's 36.7$\mu$s.  Overall, tensor compilers excel in supporting a wide range of operators, while static kernel libraries are more adept at deeply optimizing special operators. Despite this, Pruner demonstrates a stable performance advantage across the other models tested.

\begin{figure}[t]
\begin{center}
\centerline{\includegraphics[width=1\columnwidth]{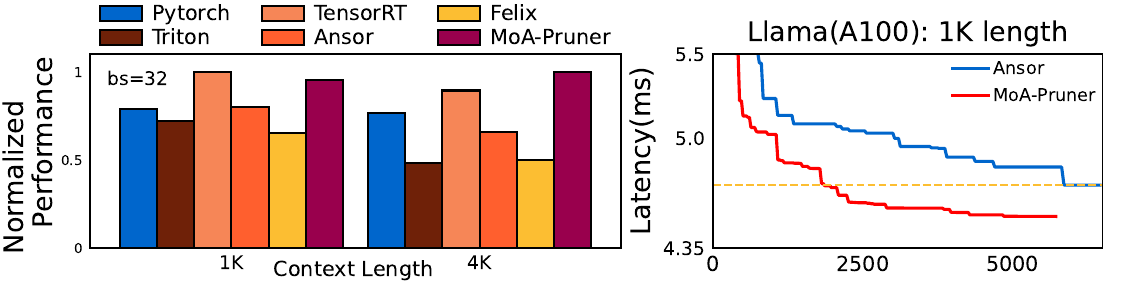}}
\caption{(left) Normalized inference performance comparison on NVIDIA A100 with context length of 1K and 4K; (right) Tuning curve for Llama with 1K context length.}
\label{large_context_fp32_comparison}
\Description{}
\end{center}
\end{figure}

\textbf{Pruner's robustness in long context LLMs decoding.} Under full-precision optimization, we selected Llama to evaluate the robustness of MoA-Pruner with input lengths of 1K and 4K and a batch size of 32. Figure \ref{large_context_fp32_comparison} presents the normalized decoding inference performance of Llama on the A100, comparing MoA-Pruner with off-the-shelf inference frameworks and tensor compilers. MoA-Pruner achieves competitive performance to the TensorRT. For other methods, it achieves speedups ranging from 1.28$\times$ for Ansor to 1.57$\times$ for Felix. Compared to other search-based compilers, MoA-Pruner consistently outperforms Ansor and Felix within the same search space. Figure \ref{large_context_fp32_comparison}  shows the tuning curve of MoA-Pruner and Ansor,  where MoA-Pruner rapidly explores the search space and maintains high-quality results.

\subsection{Single Operator Performance}
\begin{figure}[b]
\begin{center}
\centerline{\includegraphics[width=1.0\columnwidth]{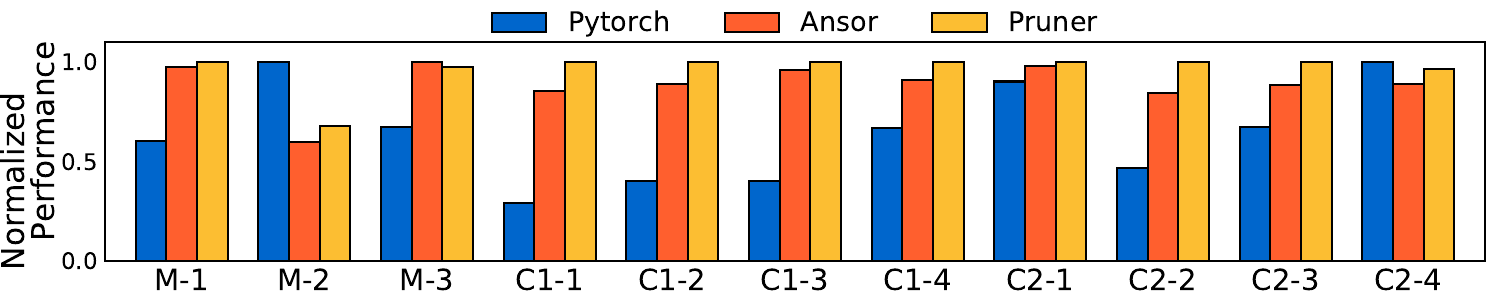}}
\caption{Normalized performance comparison with Ansor and Pytorch on NVIDIA A100, where M-k refers to $k_{th}$ matmul case, C\#-k refers to  $k_{th}$ conv2d case with stride \#.}
\label{opPerf}
\Description{}
\end{center}
\end{figure}

We evaluate the tuning performance of Pruner for some full precision operators (e.g., Matmul, Convolution) with random shapes on  A100 and compare it with Pytorch, and Ansor. We repeat the computing 500 times for each operator to obtain the average performance of Pytorch on the hardware using the nsight system. Pruner and Ansor tune each operator with 800 trials, without using pre-trained models. Figure \ref{opPerf} illustrates the comparison between different methods. Compared with Pytorch, Pruner performs exceptionally well on most operators, though it has some disadvantages on a few specific ones. The reason is that Pytorch can implement these operators through more efficient algorithms such as splitKGEMM (M-2) and Winograd \cite{lavin2016fast}. It is worth noting that Pruner achieves better performance than Ansor within a shorter search time.

\subsection{Compilation Cost}  We measured the overall compilation time and GPU memory usage in the online tuning scenarios. Table \ref{compiletime} presents the compilation times of different methods over 2,000 tuning trials across five end-to-end workloads on the NVIDIA TITAN V. For results on other platforms, refer to Figure \ref{end-to-end-tuning-curve}. The data shows that the average compilation time of Pruner and MoA-Pruner is 84.1\% and 75.3\% of Ansor's, respectively. This reduction is attributed to Pruner's "Draft-then-verify" search paradigm, which uses a draft model (LSE) for initial performance prediction, reducing the evaluation set from 8,000 candidates to 512. PaCM then verifies this subset, significantly cutting down the inference overhead. MoA further optimizes the process by lowering the training frequency. We also recorded the maximum GPU memory usage with an inference batch size of 4,096. The proposed PaCM method uses 1,694 MB of GPU memory, while TenSetMLP/Ansor uses 1,546 MB, and TLP requires 4,812 MB. Compared to existing methods, our approach only slightly increases GPU memory demand. These results demonstrate the effectiveness of our method in significantly reducing compilation time while maintaining competitive GPU memory usage.

\begin{table}[h]
\small
\centering
\caption{Compilation time (min) with tuning 2,000 trials.}
\label{compiletime}
\begin{tabular}{c|ccccc}
\toprule
 $\rm \textbf{Method}$ & R50 & I-V3 & ViT & Dl-V3  & B-base \\
\midrule
 Ansor & 124.63 & 123.15 & 99.38 & 120.4  & 117.35 \\
 Pruner & 102.03 & 96.57 & 93.47 & 100.92  & 102.95 \\
MoA-Pruner & \textbf{91.67} & \textbf{90.08} & \textbf{82.27} & \textbf{91.25}  & \textbf{89.35} \\
\bottomrule
\end{tabular}
\end{table}

\begin{figure}[b]
\begin{center}
\centerline{\includegraphics[width=1\columnwidth]{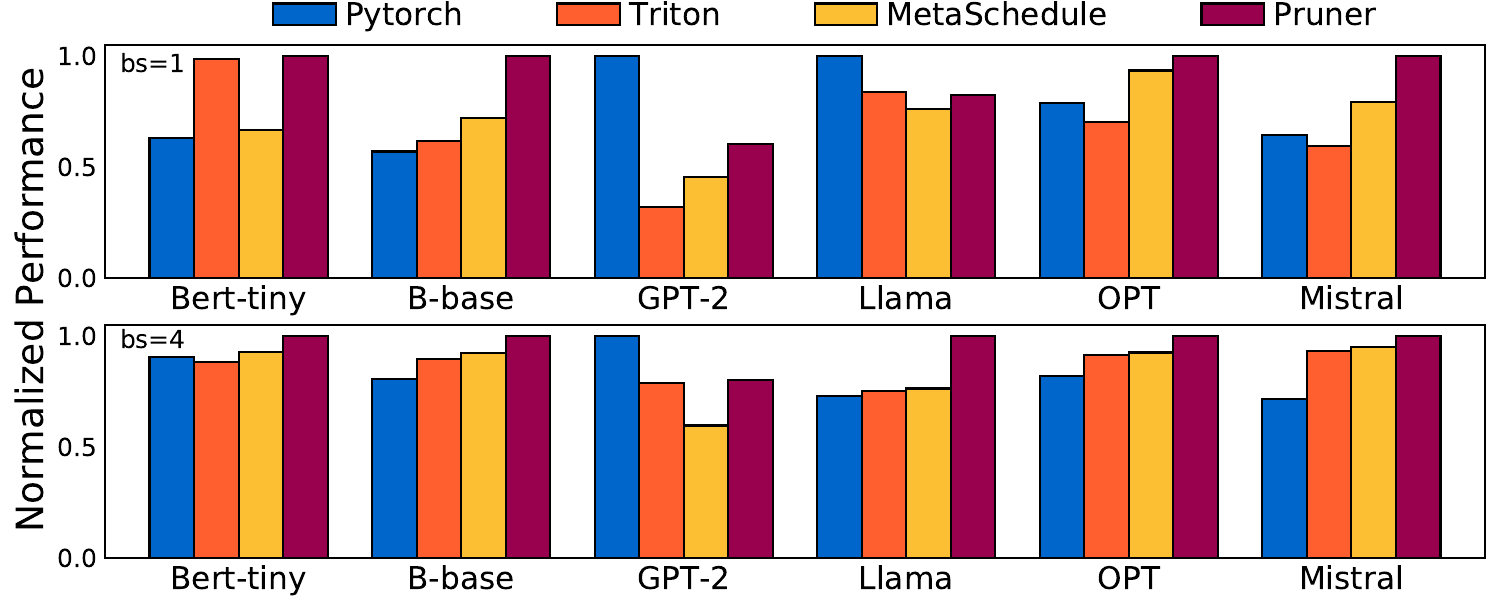}}
\caption{Normalized inference performance comparison on NVIDIA A100 TensorCore with batch sizes of 1 and 4.}
\label{tc_majorrevision}
\Description{}
\end{center}
\end{figure}

\subsection{Compile on TensorCore}

We compared Pruner to two tensorcore-supported tensor compilers: MetaSchedule (tensorcore-supported search framework in TVM), Triton, and one DNN framework: Pytorch (cudaLib). We integrate Pruner's technology into MetaSchedule by introducing a Symbol to describe TensorCore resources utilization into LSE and another new dataflow between shared and fragment into PaCM. We use 16$\times$16$\times$16 WMMA instruction in Pruner and select six Transformer-based language model variants with half-precision(H) optimization, including Bert-Tiny/Base, GPT-2, OPT, Llama, and Mistral, whose majority of ops are matrix multiplication, as our tensorcore benchmarks. The batch size and context length for each benchmark are set to (1,4) and 128, respectively, as we report in Table~\ref{dnn_a}. Figure \ref{tc_majorrevision} shows the normalized inference performance on the A100 tensor core. For MetaSchedule, Pruner achieves an average 1.22$\times$ improvement for the entire tuning performance. Pruner achieves an average $1.23\times$ and $1.3\times$ speedup against Pytorch and Triton, respectively. It is worth mentioning that in particular cases (e.g., GPT-2 and Llama), the deeply optimized and hand-tuning kernel libraries in PyTorch can outperform Pruner auto-tuning. For instance, in GPT-2, table \ref{fp16_gpt2_llama} presents the four linear operators inference latency of GPT-2 on A100 TensorCore, where batch size is 1 and prefill context length is 128. When the reduction axis dimension is relatively large (e.g., 3072), cudaLib uses splitK to optimize performance, potentially yielding superior solutions. Besides this op, Pruner consistently delivers stable and competitive results for the first three ops. However, in most general cases, Pruner achieves comparable and superior performance, avoiding hand-tuning for any unseen models and kernels. Overall, tensor compilers excel at supporting a wide variety of operators, whereas static kernel libraries are better suited for deeply optimizing specialized operators.

\begin{table}[h]
\footnotesize
\centering
\caption{Comparison of linear operator inference latency ($\mu$s) in GPT-2 on NVIDIA A100 TensorCore with batch size of 1 and prefill context length of 128.}
\label{fp16_gpt2_llama}
% \resizebox{\linewidth}{!}{
\begin{tabular}{c|cc|cc|c}
\toprule
 ID & Input Shape & Weight Shape & cudaLib & splitK & Pruner \\
\midrule
1 & (1, 128, 768)& (768, 2304) & 13.17 & w/o & \textbf{11.63} \\
2 & (1, 128, 768) & (768, 768) & 10.96 & w & \textbf{9.53}\\
3 & (1, 128, 768) & (768, 3072) & 14.01 & w/o & \textbf{12.84}  \\
4 & (1, 128, 3072) & (3072, 768) & \textbf{18.96} & w & 23.46\\
\bottomrule
\end{tabular}
% }
\end{table}

\begin{table}[b]
\footnotesize
\centering
\caption{Search speedup of Pruner, measured by how much time Pruner takes to reach the best MetaSchedule's entire search performance on  NVIDIA A100 TensorCore.}
\label{tc_trials}
\resizebox{\linewidth}{!}{
\begin{tabular}{c|cccccc}
\toprule
 $\rm \textbf{Input Shape}$ & Bert-Tiny & Bert-Base & GPT-2 & Llama  & OPT & Mistral \\
\midrule
 (1, 128) & 8.53$\times$ &  4.17$\times$ & 5.7$\times$  & 2.03$\times$ & 1.96$\times$  & 4.15$\times$  \\
 (4, 128) & 2.76$\times$ & 1.85$\times$ & 9.26$\times$ & 3.27$\times$ & 3.26$\times$  &  1.99$\times$ \\
\bottomrule
\end{tabular}
}
\end{table}

As a search-based tensor compiler, we also summarize the speedup of Pruner compared to Metaschedule in schedule search time. Table \ref{tc_trials} shows the search speedup, measured by how much time Pruner takes to reach the best performance achieved by MetaSchedule's entire search. On average, Pruner achieves a  $4.08\times$ speedup in schedule search time against MetaSchedule. Similar to results in full-precision optimization, this search speedup is primarily due to LSE's initial exploration of the space, which reduces the inference overhead of the learned cost model. Additionally, the dataflow features in PaCM make it easier to predict the schedule performance of matmul operators, enabling faster discovery of high-performance schedules.

\begin{figure}[h]
\begin{center}
\centerline{\includegraphics[width=1\columnwidth]{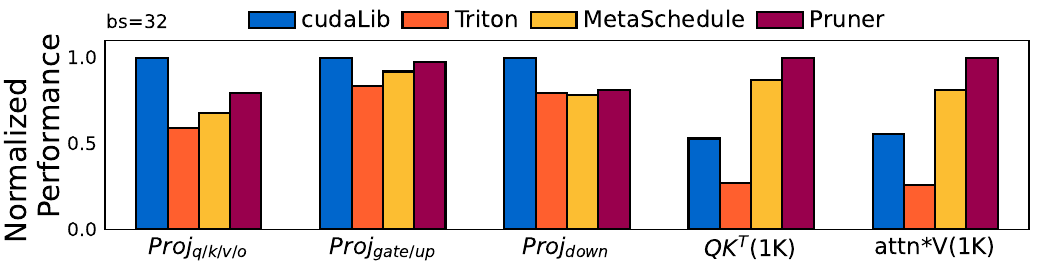}}
\caption{Normalized inference performance comparison of operators within Llama on NVIDIA A100 TensorCore.}
\label{fp16_llama_final_revision_comparision}
\Description{}
\end{center}
\end{figure}

Under half-precision optimization, we test each operator's performance during  Llama decoding with 1K context and batch size of 32. Figure \ref{fp16_llama_final_revision_comparision} shows the performance comparison of five ops within transformer blocks during decoding. Since token-by-token generation during decoding, the linear ops are fixed matrix multiplications (fixed batch size and $n=1$), while attention computations vary due to KV pairs decided by sequence length.  For linear ops,  where the dimension of the reduction axis is relatively large, cudaLib outperforms other methods due to using the splitK technique for optimization. In attention computations, the multi-heads (merged into batch dimension) increase the dimension of the parallel axis, making parallelization easier. Pruner can achieve high performance within the search space defined by simple tiling rules.  Based on the experiments, the LLM prefill phase involves diverse tensor programs due to variable prefill context lengths. As computations primarily are matrix multiplications with large parallel dimensions, search-based compilers can efficiently exploit parallelism through tiling. They are also effective for attention computations, where KVCache variations and multi-heads expand the parallel dimension. In contrast, manual optimization is better suited for these fixed linear operators (e.g., Proj layers) with relatively large reduction axis dimensions of the decoding phase.

% \BlankLine
\subsection{Dataset-based Metrics Analysis}
\label{fana}

We also use dataset-based metrics to evaluate the performance of the designed cost model on a static dataset. We verify the effectiveness of LSE and PaCM meets our design goals on the  TenSet\cite{zheng2021tenset}, which contains over 2,308 subgraphs and 4,000 schedules for each subgraph, with a total of 16 million tensor programs on NVIDIA K80 and T4. Consistent with TenSet \cite{zheng2021tenset} and TLP \cite{zhai2023tlp}, we use ResNet50, ResNet3D18, MobileNet-v2, BERT-base/tiny as the test set to validate the Tok-$k$ metrics (Eq. \ref{Topk}). Where $w_{i}$ is the appearance times of the subgraph $i$ in the model. Among all tensor programs of subgraph $i$, $L^*_{i}$ and $L_{i,j}$ are the minimum latency and the latency corresponding to the $j$-th large score of the learned cost model.

\begin{equation}
    \label{Topk}
    Top_k=\frac{\sum_{i}{L^*_{i} \times w_{i}}}{\sum_{i}{min(L_{i,j})\times w_{i}}}, 1\leq j \leq k
\end{equation}

 We also use Best-$k$ (Eq. \ref{best_k}) to evaluate  LSE, where $\hat{L}_{i,k}$ is the $k$-th best latency of $\mathcal{S}_{spec}$ generated by LSE.

\begin{equation}
    \label{best_k}
    Best_k=\frac{\sum_{i}{L^*_{i} \times w_{i}}}{\sum_{i, L^*_{i,k} \in M}{ \hat{L}_{i,k}  \times w_{i}}} 
\end{equation}

\begin{table}[h]
\footnotesize
\caption{$Best_1$ score of the $\mathcal{S}_{spec}$ with different size, where w/o refers to  remove $\mathcal{P}_{li,c}$ and $\mathcal{P}_{li,m}$ during LSE.}
\label{AblationHDonTensetGPUs}
\centering
\begin{tabular}{l|cccc}
\toprule

    \multirow{2}{*}{Method} & \multicolumn{4}{c}{size of the $\mathcal{S}_{spec}$}  \\
    \cmidrule{2-5}
    & 50 & 128 & 256 & 512 \\
\midrule
w/o $\mathcal{P}_{li,c}$ & 0.685 & 0.783 & 0.842 & 0.880 \\
w/o $\mathcal{P}_{li,m}$ & 0.757 & 0.838 & 0.886 & 0.930  \\
LSE(Ours) & 0.914 & \textbf{0.968} & \textbf{0.986} & \textbf{0.995} \\
\bottomrule
\end{tabular}
\end{table}

\begin{figure}[b]
\begin{center}
\centerline{\includegraphics[width=\columnwidth]{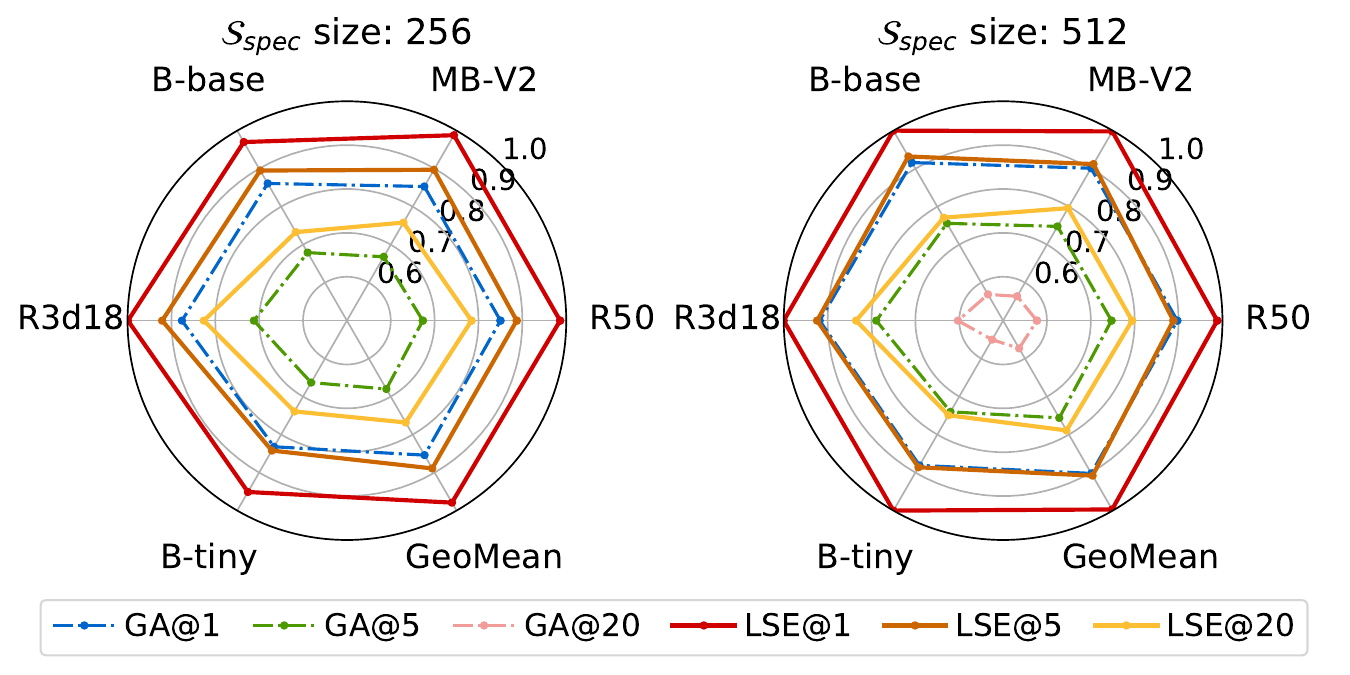}}
\caption{$Best_k$ score comparison, where the $\mathcal{S}_{spec}$  are generated by random GA and LSE on TenSet T4, respectively. @k refers to the $Best_{k}$ score of $\mathcal{S}_{spec}$.}
\label{qualityofspace}
\Description{}
\end{center}
\end{figure}

\noindent\textbf{Can Pruner draft a  high quality $\mathcal{S}_{spec}$ ?}  Based on a TenSet GPU (T4) dataset, we simulated the schedule exploration for generating $\mathcal{S}_{spec}$, representing a 4,000 exploration size for each subgraph in a DNN. We use best-$k$ to assess the quality of $\mathcal{S}_{spec}$ generated by LSE. Since those search-based DLCs rely on a learned cost model, GA employs a random search strategy with reporting an average of 5000 repeated. Figure \ref{qualityofspace} reports the best-$k$ scores of different DNNs under different sizes (256 and 512), demonstrating that LSE can explore and preserve the optimal or near-optimal schedules for each subgraph (LSE@1 is close to or equal to 1).  When the size of $\mathcal{S}_{spec}$ is reduced to 256, LSE@\# shows no significant fluctuations and maintains stable exploration quality compared to GA. We also conduct an ablation study on the core mechanisms of LSE. Table \ref{AblationHDonTensetGPUs} shows the quality degradation after removing each penalty on TenSet. Notably, $\mathcal{P}_{li,c}$ has a significant impact, showing that analysis through computational resource scheduling offers more accurate insights.

\begin{figure}[t]
\begin{center}
\centerline{\includegraphics[width=\columnwidth]{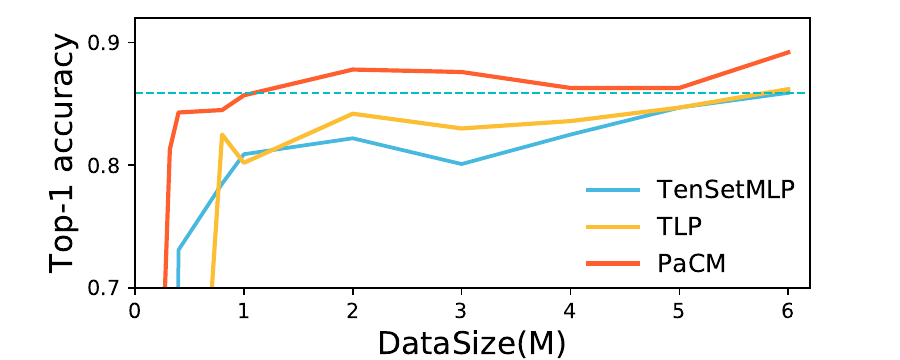}}
\caption{$Top_1$  curves comparison of PaCM, TenSetMLP, and TLP using various training data sizes.}
    \label{trainingcurveT4}
\Description{}
\end{center}
\end{figure}

\noindent\textbf{Can Pruner verify the best candidates from $\mathcal{S}_{draft}$?} We aim to design easy-training features and achieve high-precision prediction through the attention branch in PaCM. To verify whether the attention branch of PaCM meets our design goals, we conducted the following experiments. We compared the prediction accuracy of PaCM with existing deep learning-based cost models, including TenSetMLP and TLP. For each hardware configuration, we sampled 6 million data from the TenSet dataset and trained these different cost models under the same experimental setup. Table \ref{tensetgpucompare} presents the $Top_k$ prediction accuracy scores of the various cost models on the TenSet test set. The results show that PaCM significantly outperforms both TLP and TenSetMLP. Furthermore,  we also recorded the $Top_1$ score curves of different methods under various training dataset sizes. Figure \ref{trainingcurveT4} illustrates that PaCM achieves better convergence across different data scales and surpasses other fully trained models with only a small amount of training data. This demonstrates that the designed temporal dataflow features facilitate effective training of Transformer-based cost models.

\begin{table}[h]
\footnotesize
\centering
\caption{$Top_k$ score comparison of different methods on TenSet GPUs dataset.}
\label{tensetgpucompare}
\begin{tabular}{c|cc|cc}
\toprule
\multirow{2}{*}{Method} & \multicolumn{2}{c}{TenSet T4} & \multicolumn{2}{c}{TenSet K80} \\
    \cmidrule{2-5}
    & top-1 & top-5 & top-1 & top-5\\
\midrule
TenSetMLP & 0.859 & 0.941 & 0.878 & 0.958 \\
TLP  & 0.862 & 0.935 & 0.880 & 0.947  \\
PaCM(ours) & \textbf{0.892}& \textbf{0.962} &  \textbf{0.897} & \textbf{0.969}\\
\bottomrule
\end{tabular}
\end{table}

\begin{figure}[t]
\begin{center}
\centerline{\includegraphics[width=0.95\columnwidth]{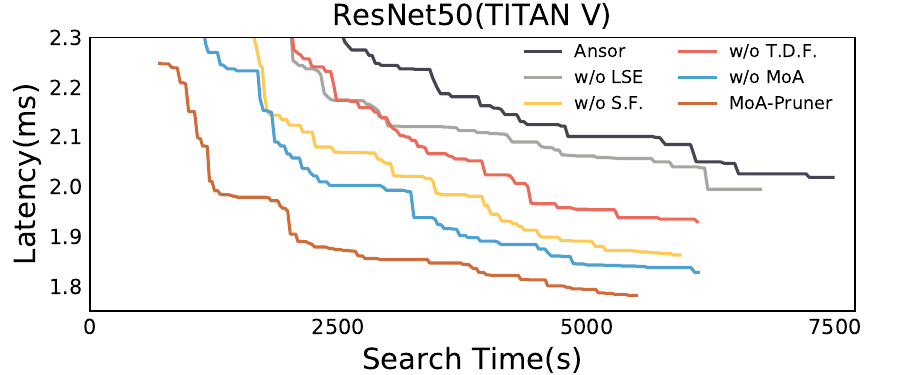}}

    \caption{Ablation study of ResNet50's tuning curve.}
    \label{R50_ab}
\Description{}
\end{center}
\end{figure}

\begin{table}[b]
\footnotesize

\caption{Ablation study of Pruner's online tuning mode.  S.F. and T.D.F represent the statement and temporal dataflow features, respectively. O-F refers to Pruner using online finetuning. The value means inference latency (ms).}
\label{AblationAI}
% \resizebox{\linewidth}{!}{
\centering
\begin{tabular}{l|ccccc}
\toprule
Method & R50 & I-V3 & ViT & Dl-V3 & B-tiny \\
\midrule
Ansor & 2.019 & 4.153 & 2.77 & 4.168 & 1.767 \\
w/o LSE & 1.995 & 3.990 &  2.562 &  4.136	 &  1.670 \\
w/o S.F. & 1.863 & 3.947 & 2.519 & 3.890  &  1.703 \\
w/o T.D.F & 1.930 & 3.898 & 2.771 & 3.838 &  1.704\\
w/o MoA & 1.828 & \textbf{3.856} & 2.489 & 3.854 & \textbf{1.611} \\
w/ O-F & 1.812 & 3.902 & 2.451 & 3.844 & 1.693 \\
MoA-Pruner & \textbf{1.782} & 3.866 & \textbf{2.432} & \textbf{3.713} & 1.655 \\
\bottomrule
\end{tabular}
% }
\end{table}

\begin{table}[b]
\footnotesize
\centering
\caption{Ablation study of Pruner's offline tuning mode performance, where perf and cost refer to inference latency (ms) and compilation cost (min).}
\label{abstudyonoffline}
\begin{tabular}{c|cc|cc}
\toprule
 \multirow{2}{*}{Models} & \multicolumn{2}{c}{w/o LSE} & \multicolumn{2}{c}{Offline Pruner} \\
    \cmidrule{2-5}
  & perf & cost & perf & cost\\
\midrule
 ResNet50 &  1.491 & 111 & \textbf{1.444}  &  \textbf{89}\\
 Inception-v3 & 2.831 & 113 & \textbf{2.687}  & \textbf{91} \\
 Bert-Base  & 3.88 & 115 & \textbf{3.639}  & \textbf{96}\\
Bert-Tiny & 1.432 & 112 & \textbf{1.326}  & \textbf{91} \\
\bottomrule
\end{tabular}
\end{table}

\subsection{Ablation Study}
We performed an ablation study on key components of MoA-Pruner, where we separately removed LSE, MoA, and the individual feature embedding branches of PaCM. Table \ref{AblationAI} shows the tuning latency across various configurations for a subset of workloads, while Figure \ref{R50_ab} illustrates the tuning curves for ResNet50 on TITAN V. We find that removing any component will increase tuning latency and compilation cost, with the most significant latency increase occurring when LSE is removed. The reason is that without LSE's initial space exploration, the learned cost model is hard to find the best candidates from the search space, rapidly, and increases the overhead of the learned cost model. Moreover, eliminating temporal dataflow features has a more negative impact than removing statement-level features, as temporal dataflow features provide a more easily trainable feature branch, making them crucial for the performance of the tuning process. Additionally, we found that the MoA strategy outperforms traditional online fine-tuning. This is because the limited and biased data collected early in online training can disrupt model optimization, whereas MoA mitigates this issue with a momentum-based approach.

In the context of offline tuning, where we have a well-trained cost model (e.g., well-pre-trained on NVIDIA A100), we also conducted a study to determine whether LSE is still necessary for the initial exploration of the search space. Table \ref{abstudyonoffline} shows that using LSE still results in faster search times. This is because the inference of the DNN-based cost model involves complex feature extraction and requires GPU resources, resulting in a much higher computational overhead than LSE. In contrast, LSE not only significantly reduces the inference times of the cost model, but also guarantees good tuning performance.

\section{Related Work}
\textbf{Automatic tuners and cost models}.
Recently, numerous compiler projects have given rise to various schemes, many of which are based on open-source efforts such as Halide \cite{ragan2013halide}, TACO \cite{kjolstad2017tensor}, XLA \cite{xla}, AKG \cite{zhao2021akg}, TVM \cite{chen2018tvm}, and nnFusion \cite{ma2020rammer}. Among them,  AutoTVM \cite{chen2018learning}, the first search framework integrated into TVM, innovatively treats the tensor program optimization as the scheduling primitive optimization and search within a manually defined template search space. Adatune \cite{li2020adatune} introduces an adaptive evaluation method that statistically early terminates a costly hardware measurement. To further improve optimization quality, FlexTensor \cite{zheng2020flextensor} and Ansor \cite{zheng2020ansor} realize the automatic generation of search space, mitigating the efficiency drawbacks of manual design. Bolt \cite{xing2022bolt}  uses hardware-native templated search to bridge the performance gap between tensor compilers and hardware-native libraries. TIRAMISU \cite{baghdadi2019tiramisu} and AKG \cite{zhao2021akg} explore using polyhedral optimization technology to search for optimal solutions. MetaSchedule \cite{shao2022tensor} supports automatic sketch generation for new special hardware. Roller \cite{zhu2022roller} can derive a candidate set of tensor programs from empirical formulas, relying on accurate hardware modeling, and requires specific constraints on each operator. Heron \cite{bi2023heron} designs hardware-specific constraint rules and a corresponding constraint-based genetic algorithm to explore search space. TenSetMLP \cite{zheng2021tenset} and TLP \cite{zhai2023tlp} extracts features from low-level code representations and high-level scheduling primitives, respectively, and adopt Multi-Layer Perceptron (MLP) and Transformer-based\cite{vaswani2017attention} models to model the mapping from features to performance.  Felix \cite{zhao2024felix}  creates a differentiable space of tensor programs, allowing efficient search of program candidates using gradient descent. TLM\cite{zhai2024enabling} introduces a language model to assist tensor program tuning.

\noindent\textbf{Cross-platform transfer strategy.}  There are few studies on cost models across hardware platforms. TenSet builds a local model to predict the gap between the source model and target hardware. Moses[37] uses model distillation to distill out transferable and non-transferable parameters.  TLP uses multi-task learning to train a multi-head cost model to predict the corresponding target performance.

\section{Conclusion}
In this paper, we propose Pruner and MoA-Pruner.  Pruner is a schedule exploration mechanism that accelerates the search process using a "Draft-then-Verify"  paradigm, including rapid exploration with a draft model and then using a more accurate learned cost model for identification from small-scale potential candidates.  MoA-Pruner introduces the momentum online adaptation to solve the pre-trained cost model cross-platform online unawareness,  enabling efficient online adaptation to any platform in online cost model tuning scenarios. Our analysis highlights the advancements and feasibility of Pruner by comparing it with existing state-of-the-art methods on three GPU-based platforms including cuda core and tensor core. Comprehensive experiments show that Pruner significantly outperforms these methods by a large margin, demonstrating its effectiveness and a commendable balance between tuning quality and efficiency. We implement Pruner in the TVM search framework (e.g., Ansor and MetaSchedule), and believe that the main idea behind Pruner complements existing search-based approaches and can be easily implemented on top of others.

\nocite{li2021analytical}

\begin{acks}
We thank our shepherd Mangpo Phothilimthana and  anonymous reviewers for their constructive comments. This work was supported by the Strategic Priority Research Program of Chinese Academy of Sciences (Grant No.XDB0500102), Laoshan Laboratory (No.LSKJ202300305). 
\end{acks}

\bibliographystyle{ACM-Reference-Format}
\balance
\bibliography{reference}

\end{document}